\documentclass{article}

\usepackage[preprint,nonatbib]{neurips_2020}

\usepackage[utf8]{inputenc} 
\usepackage[T1]{fontenc}    
\usepackage{hyperref}       
\usepackage{url}            
\usepackage{booktabs}       
\usepackage{amsfonts}       
\usepackage{nicefrac}       
\usepackage{microtype}      
\usepackage{amsmath,amssymb}
\usepackage{graphicx}
\usepackage{xcolor}
\usepackage{tabularx}
\usepackage{algorithm}
\usepackage[noend]{algpseudocode}
\usepackage{float}
\usepackage{wrapfig}
\usepackage{multicol}
\usepackage{enumitem}
\usepackage{cite}  

\newcommand\etc{etc\@ifnextchar.{}{.\@}}

\newcommand\ie{i.e., \@}
\newcommand\vs{vs. \@}
\newcommand\wrt{w.r.t. \@}
\definecolor{BPTT}{RGB}{134, 16, 1}
\definecolor{CBPTT}{RGB}{241, 133, 0}
\definecolor{CRBP}{RGB}{22, 79, 134}
\definecolor{RBP}{RGB}{183, 76, 187}

\newcommand{\rb}{\right]}
\newcommand{\lb}{\left[}

\title{Stable and expressive recurrent vision models}

\author{%
  Drew Linsley$^*$, \hfill Alekh K Ashok$^*$, \hfill Lakshmi N Govindarajan$^*$, \hfill Rex Liu, \hfill Thomas Serre\\
  Carney Institute for Brain Science \\
  Department of Cognitive Linguistic \& Psychological Sciences\\
  Brown University \\
  Providence, RI 02912 \\
  \texttt{\{drew\_linsley,alekh\_ashok,lakshmi\_govindarajan,} \\
  \texttt{rex\_liu,thomas\_serre\}@brown.edu} \\
}

\begin{document}

\maketitle

\begin{abstract}
Primate vision depends on recurrent processing for reliable perception~\cite{Roelfsema2000-op,Ullman1984-gz,Gilbert2013-hb}. A growing body of literature also suggests that recurrent connections improve the learning efficiency and generalization of vision models on classic computer vision challenges. Why then, are current large-scale challenges dominated by feedforward networks? We posit that the effectiveness of recurrent vision models is bottlenecked by the standard algorithm used for training them, ``back-propagation through time'' (BPTT), which has $\mathcal{O}(N)$ memory-complexity for training an $N$ step model. Thus, recurrent vision model design is bounded by memory constraints, forcing a choice between rivaling the enormous capacity of leading feedforward models \emph{or} trying to compensate for this deficit through granular and complex dynamics. Here, we develop a new learning algorithm, ``contractor recurrent back-propagation'' (C-RBP), which alleviates these issues by achieving constant $\mathcal{O}(1)$ memory-complexity with steps of recurrent processing. We demonstrate that recurrent vision models trained with C-RBP can detect long-range spatial dependencies in a synthetic contour tracing task that BPTT-trained models cannot. We further show that recurrent vision models trained with C-RBP to solve the large-scale \textit{Panoptic Segmentation} MS-COCO challenge outperform the leading feedforward approach, with fewer free parameters. C-RBP is a general-purpose learning algorithm for any application that can benefit from expansive recurrent dynamics. Code and data are available at \url{https://github.com/c-rbp}.
\end{abstract}


\section{Introduction}
\footnotetext[1]{These authors contributed equally to this work.}
\vspace{-2mm}
Ullman (1984) famously theorized that humans reason about the visual world by composing sequences of elemental computations into ``visual routines''~\cite{Ullman1984-gz}. It has been found that many of these visual routines, from perceptual grouping~\cite{Roelfsema2006-au} to object categorization~\cite{DiCarlo2007-zx}, depend on local and long-range recurrent circuits of the visual cortex~\cite{Gilbert2013-hb,Roelfsema2000-op,Kreiman2020-jm}. Convolutional neural networks (CNNs) with recurrent connections -- recurrent CNNs -- also seem to learn visual routines that standard feedforward CNNs do not~\cite{Linsley2018-ls,Kim2020-yw,Liang2015-sf,Kim2016-pg,Li2018-gx,Linsley2018-wx,Tang2018-we,Lotter2016-qr}. For example, consider the \textit{Pathfinder} challenge in Fig.~\ref{fig:state_space}a, which asks observers to trace the contour extending from the white dot. Although \textit{Pathfinder} is visually simple, clutter and variations in path shape make it difficult for feedforward CNNs to solve, even very deep residual networks~\cite{Linsley2018-ls,Kim2020-yw}. By contrast, a one-layer recurrent CNN can learn to solve \textit{Pathfinder} by incrementally grouping paths from one end to the other, reminiscent of Gestalt-like visual routines used by human observers~\cite{Houtkamp2010-wi,Linsley2018-ls,Kim2020-yw}. Others have found that the visual routines learned by recurrent CNNs on small computer vision datasets lead to better sample efficiency and out-of-distribution generalization than feedforward CNNs~\cite{Linsley2018-rc,Linsley2020-en,Kreiman2020-jm}. There is also evidence that primate visual decisions and neural responses elicited by natural images are best explained by recurrent CNNs~\cite{Kubilius2018-gd,Kubilius2019-qr,Nayebi2018-eg,Kar2019-ye,Sinz2018-jp,Kietzmann2019-do,Linsley2020-en}. Nevertheless, the great promise of recurrent CNNs has yet to translate into improvements on large-scale computer vision challenges like MS-COCO~\cite{Lin2014-zk}, which are dominated by feedforward CNNs. 


A well known limitation of recurrent CNNs is a memory bottleneck imposed by the standard learning algorithm, ``back-propagation through time'' (BPTT;~\cite{Werbos1990-fm}). The memory requirement of BPTT-trained models scales linearly with steps of processing, since optimization involves propagating error through the full latent trajectory. This makes it difficult to develop recurrent CNNs that can rival the massive capacity of leading feedforward CNNs, which is critical for performance on challenges~\cite{Tan2019-uh}, while also simulating enough steps of processing to learn robust human-like visual routines.

\paragraph{Contributions.}
We develop a solution to the recurrent CNN memory bottleneck introduced by BPTT. Our work is inspired by recent successful efforts in memory-efficient approximations to BPTT for sequence modeling~\cite{Liao2018-tl,Bai2019-wd}. Of particular interest is recurrent back-propagation (RBP), which exploits the stability of convergent dynamical systems to achieve constant memory complexity \wrt steps of processing~\cite{Almeida1987-me,Pineda1987-im,Liao2018-tl}. This approach depends on models with stable dynamics that converge to a task-optimized steady state. However, we find that leading recurrent CNNs violate this assumption and ``forget'' task information as they approach steady state. While this pathology can be mitigated with hyperparameters that guarantee stability, these choices hurt model performance, or ``expressivity''. Thus, we make the observation that recurrent CNNs face a fundamental trade-off between stable dynamics and model expressivity that must be addressed before they can adopt efficient learning algorithms and compete on large-scale computer vision challenges. 
\begin{itemize}[leftmargin=*]  
    \item We derive a constraint for training recurrent CNNs to become both stable \textit{and} expressive. We refer to this as the \textit{Lipschitz-Constant Penalty} (LCP). 
    \item We combine LCP with RBP to introduce ``contractor-RBP'' (C-RBP), a learning algorithm for recurrent CNNs with constant memory complexity \wrt steps of processing.
    \item  Recurrent CNNs trained with C-RBP learn difficult versions of \textit{Pathfinder} that BPTT-trained models cannot due to memory constraints, generalize better to out-of-distribution exemplars, and need a fraction of the parameters of BPTT-trained models to reach high performance.
    \item C-RBP alleviates the memory bottleneck faced by recurrent CNNs on large-scale computer vision challenges. Our C-RBP trained recurrent model outperforms the leading feedforward approach to the MS-COCO Panoptic Segmentation challenge with nearly 800K fewer parameters, and without exceeding the memory capacity of a standard NVIDIA Titan X GPU.
\end{itemize}

\section{Background}\label{sec:background}
We begin with a general formulation of the recurrent update step at $t \in \left\{ 1..N \right\}$ in an arbitrary layer of a recurrent CNN, which processes images with height $H$ and width $W$.
\begin{equation}
h_{t+1} = F(x, h_t, w_F).\label{eq:transition}
\end{equation}
This describes the evolution of the hidden state $h \in \mathbb{R}^{H \times W \times C}$ through the update function $F$ (a recurrent layer) with convolutional kernels $w_F \in \mathbb{R}^{S \times S \times C \times C}$, where $C$ is the number of feature channels and $S$ is the kernel size. Dynamics are also influenced by a constant drive $x \in \mathbb{R}^{H \times W \times C}$, which in typical settings is taken from a preceding convolutional layer. The final hidden state activity is either passed to the next layer in the model hierarchy, or fed into an output function to make a task prediction. The standard learning algorithm for optimizing parameters $w_F$ \wrt a loss is BPTT, an extension of back-propagation to recurrent networks. BPTT is implemented by replicating the dynamical system in Eq.~\ref{eq:transition} and accumulating its gradients over $N$ steps (SI Eq.~7). BPTT computes gradients by storing each $h_t$ in memory during the forward pass, which leads to a memory footprint that increases linearly with steps.

\paragraph{Steady state dependent learning rules} There are alternatives to BPTT that derive better memory efficiency from strong constraints on model dynamics. One successful example is recurrent back-propagation (RBP), which optimizes parameters to achieve steady-state dynamics that are invariant to slight perturbations of input representations; a normative goal that echoes the classic Hopfield network~\cite{Hopfield1982-cz}. When used with models that pass its test of stability, which we detail below, RBP memory complexity is \emph{constant} and does not scale with steps of processing (the precision of dynamics). RBP is especially effective when a system's steady states can be characterized by determining its Lyapunov function~\cite{Liao2018-tl}. However, such analyses are generally difficult for non-convex optimizations, and not tractable for the complex loss landscapes of CNNs developed for large-scale computer vision challenges. RBP assumes that dynamics of the transition function $F(\cdot)$ eventually reach an
equilibrium $h^*$  as $t \to \infty$ (Eq.~\ref{eq:steady}).\vspace{-3mm}
\begin{tabularx}{\linewidth}{XX}
\begin{equation}
h^* = F(x, h^*, w_F) \label{eq:steady}
\end{equation}
&
\begin{equation}
\Psi(w_F, h) = h - F(x,h,w_F) \label{eq:psiFn}
\end{equation}
\end{tabularx}\vspace{-3mm}
In other words, $h^*$ is unchanged by additional processing. We can construct a function $\Psi(\cdot)$ such that the equilibrium $h^*$ becomes its root (Eq.~\ref{eq:psiFn}); i.e. $\Psi(w_F, h^*)=0$. RBP leverages the observation that when differentiating Eq.~\ref{eq:psiFn}, the gradient of steady state activity $h^*$ \wrt the parameters of a stable dynamical system $w_F$ can be directly computed with the Implicit Function Theorem~\cite{Rudin1964-nr,Pineda1987-im, Almeida1987-me,Scarselli2009-vs}.
\begin{equation}
\frac{\partial h^*}{\partial w_F} = \left(I - J_{F, h^*} \right)^{-1} \frac{\partial F(x,h^*,w_F)}{\partial w_F}
\end{equation}
Here, $J_{F, h^*}$ is the Jacobian matrix $\partial F(x,h^*,w_F) / \partial h^*$ (SI \S3). In practice, the matrix $\left(I - J_{F, h^*} \right)^{-1}$ is numerically approximated~\cite{Almeida1987-me,Liao2018-tl}. RBP is designed for RNNs that pass its \textit{constraint-qualifications} test for stability, where \textbf{(i)} $\Psi(\cdot)$ is continuously differentiable with respect to $w_F$ and \textbf{(ii)} $\left(I - J_{F, h^*} \right)$ is invertible. When these conditions are satisfied, as is often the case with neural networks for sequence modeling, RBP is efficient to compute and rivals the performance of BPTT~\cite{Liao2018-tl}.
\subsection{Do recurrent CNNs pass the {constraint-qualifications} test of RBP?}\label{sec:pf_methods}
\begin{figure}[t]
\begin{center}
  \includegraphics[width=1\linewidth]{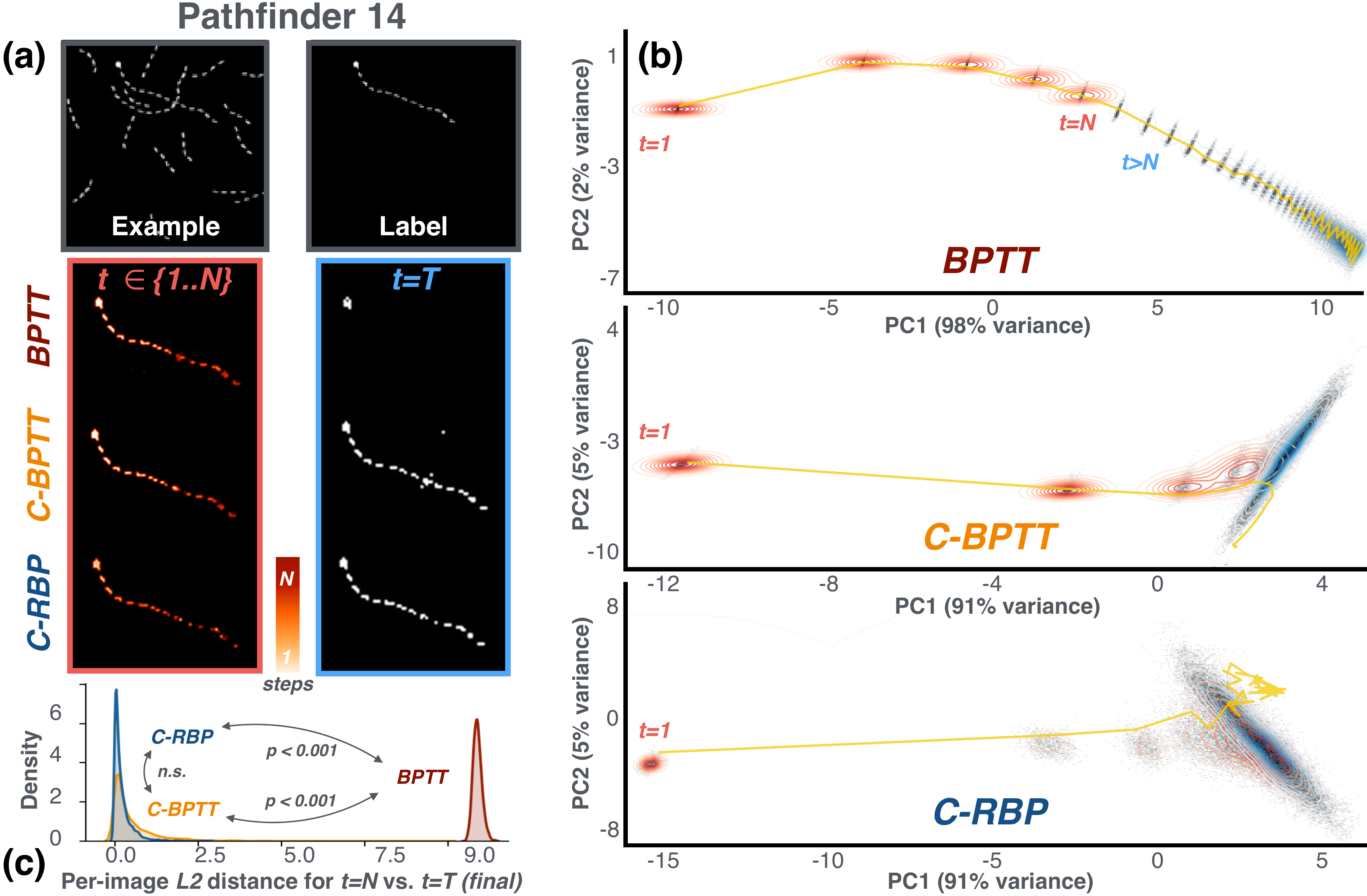} \caption{Recurrent CNNs trained with backpropagation through time (\textcolor{BPTT}{BPTT}) have unstable dynamics and forget task information. This pathology is corrected by our \textit{Lipschitz Coefficient Penalty} (LCP). \textbf{(a)} Incremental segmentations of horizontal gated unit (hGRU) models trained for $N$ recurrent steps on Pathfinder-14. Heatmaps in the left column show the models tracing the target contour until $t\text{=}N\text{=}6$ steps. The right column depicts model predictions after $t\text{=}T\text{=}40$ steps. Segmentations from the \textcolor{BPTT}{BPTT}-trained hGRU degenerate but models trained with LCP did not. Learning algorithms of LCP-trained models are ``\textcolor{CBPTT}{contractor-BPTT}'' (\textcolor{CBPTT}{C-BPTT}) and ``\textcolor{CRBP}{contractor-RBP}'' (\textcolor{CRBP}{C-RBP}). \textbf{(b)} Visualization of horizontal gated unit (hGRU) state spaces by projecting hidden states onto each model's top-two eigenvectors. Grey dots are the 2D-histogram of projected hidden states, red contours are hidden state densities up to the task-optimized $N$ steps, and blue contours are hidden state densities following that step. Dynamics for an exemplar image are plotted in yellow. \textcolor{BPTT}{BPTT}-trained model dynamics diverge when $t>N$, but models trained with LCP did not. \textbf{(c)} Two-sample KS-tests indicate that the distance in state space between $t=N$ and $t=T$ hidden states is significantly greater for an hGRU trained with \textcolor{BPTT}{BPTT} than an hGRU trained with \textcolor{CBPTT}{C-BPTT} or \textcolor{CRBP}{C-RBP} (n.s.$\ =\ $ not significant).}\label{fig:state_space}\vspace{-4mm}
\end{center}\end{figure}

We turn to the \textit{Pathfinder} challenge \cite{Linsley2018-ls,Kim2020-yw} to test whether RBP can optimize recurrent CNN architectures devised for computer vision (see \ref{fig:state_space}a and Fig.~S2 for examples). \textit{Pathfinder} is an ideal test bed for recurrent vision models because they can more efficiently solve it than feedforward models.
\begin{itemize}[leftmargin=*]  
\item \textit{Pathfinder} tests the ability of models to detect long-range spatial dependencies between features -- identifying the target contour and tracing it from one end to the other. Feedforward architectures (like ResNets) need a sufficient amount of depth to learn such dependencies, leading to an explosion of parameters and learning problems~\cite{Linsley2018-ls,Kim2020-yw}. In contrast, recurrent CNNs can broaden receptive fields over steps of processing without additional processing layers (SI \S3), but doing so requires maintaining task information across dynamics.

\item \textit{Pathfinder} is parameterized for fine-grained control over task difficulty. By lengthening or shortening target contours and clutter, we can generate more or less difficult datasets (Fig.~S2) while controlling other perceptual variables that could introduce spurious image/label correlations.
\end{itemize}

In summary, recurrent CNNs that can solve \textit{Pathfinder} need to learn a dynamic visual routine for detecting long-range spatial dependencies in images. BPTT-trained recurrent CNNs can do this~\cite{Linsley2018-ls,Kim2020-yw}. Here we test whether RBP-trained models can do the same.

\paragraph{Methods}We trained the leading recurrent CNN for \textit{Pathfinder}, the horizontal gated recurrent unit (hGRU; a complete model description is in SI \S3.1), to solve a version where the target contour is 14-dashes long (Fig.~\ref{fig:state_space}a). We modified \textit{Pathfinder} from its origins as a classification task~\cite{Linsley2018-wx} to a segmentation task, which makes it easier to interpret model dynamics and translate our findings to the large-scale MS-COCO challenge examined in \S\ref{sec:COCO}. The hGRU architecture consisted of \textbf{(i)} an input layer with 24 Gabor-initialized convolutional filters and one difference-of-Gaussians filter, followed by \textbf{(ii)} an hGRU with $15\times15$ horizontal kernels and 25 output channels, and finally \textbf{(iii)} a 1$\times$1 convolutional ``readout'' that transformed the final hGRU hidden-state to a per-pixel prediction via batch normalization~\cite{Ioffe2015-zm} and a $1\times1$ convolutional kernel. We began by testing four versions of the hGRU: one trained with BPTT for 6 steps, which was the most that could fit into the 12GB memory of the NVIDIA Titan X GPUs used for this experiment; and versions trained with RBP for 6, 20, and 30 steps. We also trained a fifth control model, a feedforward version of the 6 step BPTT-trained hGRU, where parameters were not shared across time (``feedforward CNN control''). The models were trained with Adam~\cite{Kingma2014-ct} and a learning rate of $3\text{e-}4$ to minimize average per-pixel cross entropy on batches of 32 images. Training lasted 20 epochs on a dataset of 200,000 images. Performance was measured after every epoch as the mean intersection over union (\textit{IoU}) on a held-out test set of 10,000 images. We report the maximum \textit{IoU} of each model on this test set.
\paragraph{Results} The three hGRUs trained with RBP performed poorly on Pathfinder-14 (6 step: 0.50 \textit{IoU}; 20 step: 0.71 \textit{IoU}; 30 step: 0.70 \textit{IoU}), far worse than a BPTT-trained hGRU (0.98 \textit{IoU}). The RBP-hGRUs were also outperformed by the feedforward CNN control (0.78 \textit{IoU}), although this control used 3 times more parameters than the hGRUs.

Why does an hGRU trained with RBP fail to learn \textit{Pathfinder}? To address this question, we return to the {constraint-qualifications} test of RBP. The hGRU, like all models successfully optimized with gradient descent, satisfies condition (\textbf{i}) of the test: it is composed of a series of differentiable functions. This means that the problem likely arises from condition (\textbf{ii}), which requires the matrix $I-J_{F,h^*}$ to be invertible. Indeed, if $I-J_{F,h^*}$ is singular and not invertible, training dynamics will devolve into an unstable regime as we observed for RBP-hGRU training (Fig.~S7).
One way of guaranteeing an invertible $I-J_{F,h^*}$ is by forcing $F$ to be a contractive map~\cite{Scarselli2009-vs}. When this constraint is in place, we can invoke Banach fixed point theorem to ensure convergence to a unique fixed point starting from any initial hidden state $h_0$. To elaborate, $F$ implements a contraction map and will converge onto a unique fixed point if the following holds: $\forall h_i, h_j, x$, where $i,j \in \mathbb{R}_+$, $\exists \lambda\in [0,1)$ such that
\begin{equation}
\left\lVert F(h_i; x, w_F) - F(h_j; x, w_F) \right\rVert_2  \leq \lambda \left\lVert h_i - h_j\right\rVert_2\label{eq:bound}
\end{equation}
A straightforward way of forcing a contractive map is by building models with hyperparameters that are globally contractive (\ie across the entire latent trajectory of dynamics), like squashing non-linearities (\textit{sigmoid} and \textit{tanh}~\cite{Glorot2011-la,Glorot2010-fa}). However, these same hyperparameters are suboptimal for computer vision, where unbounded non-linearities (like \textit{ReLU}) are used because of their control over vanishing gradients in deep hierarchical networks and improving function expressivity~\cite{Gulcehre2016-tk,Clevert2015-bz}. 

In principle, recurrent CNNs like convolutional LSTMs, that use \textit{tanh} non-linearities, are better suited for RBP optimization than an hGRU, which uses soft unbounded rectifications (\textit{softplus}). Indeed, we found that a 20-step convolutional LSTM (``convLSTM''; architecture detailed in SI \S3.2) trained with RBP on Pathfinder-14 performs slightly better (0.73 \textit{IoU}) than the RBP-trained hGRU (0.71 \textit{IoU}). At the same time, the RBP-trained convLSTM performs much worse than a 6-step BPTT-trained convLSTM (0.81 \textit{IoU}) and a BPTT-trained hGRU (0.98 \textit{IoU}). In other words, the convLSTM is more stable but less expressive than the hGRU. Furthermore, an hGRU with squashing non-linearities could not be trained reliably with RBP due to vanishing gradients. These findings raise the possibility that recurrent CNNs face a trade-off between ``expressivity'' needed for competing on large-scale computer vision challenges, and ``stability'' that is essential for using learning algorithms like RBP which rely on equilibrium dynamics.

\paragraph{Expressivity \vs stability} To better understand the trade-off faced by recurrent CNNs, we examined the stability of an \emph{expressive} recurrent CNN: the BPTT-trained hGRU, which outperformed all other architectures on Pathfinder-14. The hGRU solves Pathfinder by incrementally grouping the target contour, making it possible to visualize the evolution of its segmentation over time. By passing the model's hidden states through its readout, we observed that task information vanishes from its hidden states after the task-optimized $N$-steps of processing (Fig.~\ref{fig:state_space}a; compare predictions at $t\text{=}N$ and $t\text{=}T$). These unstable dynamics were not specific to the BPTT-trained hGRU. We found similar results for a BPTT-trained convLSTM (Fig.~S3), and when optimizing hGRUs with common alternatives to BPTT (Fig.~S4). Next, we performed a state space analysis to measure model dynamics on all images in the Pathfinder-14 test set (method described in SI \S3.3). The state space revealed a large divergence between hGRU hidden state activity at the task-optimized $t\text{=}6\text{=}N$ step \vs activity near steady state at $t\text{=}40\text{=}T$ (Fig.~\ref{fig:state_space}b). There was nearly as large of a difference between hidden states at $t\text{=}1$ and $t\text{=}N$ as there was between hidden states at $t\text{=}N$ and $t\text{=}T$. 

\section{Stabilizing expressive recurrent CNNs}
While RNN stability is not necessary for all applications~\cite{Miller2018-bu}, it is critical for constant-memory alternatives to BPTT like RBP, which we hypothesized would improve recurrent CNN performance on large-scale vision challenges. Thus, we designed a ``soft'' architecture-agnostic constraint for learning local contraction maps, which balances model expressivity and stability over the course of training. Our goal was to derive a constraint to keep the largest singular value of $J_{F, h^*}<1$, and force $F$ to be locally contractive at $h^*$ (SI \S1.2; this contrasts with the global contraction across dynamics enforced by squashing non-linearities, which is problematic in computer vision as demonstrated in \S\ref{sec:pf_methods}). However, the size of this Jacobian is quadratic in hidden state dimensions, making it too large to feasibly compute for recurrent CNNs. We overcome this limitation by introducing an approximation, our \textit{Lipschitz Coefficient Penalty} (LCP), which constrains $\left( \mathbf{1} \cdot J_{F, h^*} \right)_i < \lambda \; \forall i$, where $i$ is a column index.
\begin{equation}
\lVert( \mathbf{1} \cdot  J_{F, h^*} - \lambda)^+\rVert_2\label{eq:LCP}
\end{equation}
Here, $(\cdot)^+$ denotes element-wise rectification and $\lambda \in [0,1)$ is the hand-selected Lipschitz constant which bounds $\lVert J_{F, h^*} \lVert_2$ and hence the degree of contraction in $F$. The LCP (Eq.~\ref{eq:LCP}) can be combined with any task loss for model optimization, and circumvents the need to explicitly determine the Jacobian $J_{F,h^*}$. We show in SI \S1.2 that the vector-Jacobian product used in the LCP, which is efficiently computed through \textit{autograd} in deep learning libraries, serves as a approximate upper bound on the spectral radius of the $J_{F, h^*}$. Note that because minimizing the LCP implicitly involves the Hessian of $F$, it is best suited for recurrent CNNs that use activation functions with defined second derivatives (like \textit{softplus} for the hGRU).

\begin{figure}[t]
\begin{center}  
  \includegraphics[width=1\linewidth]{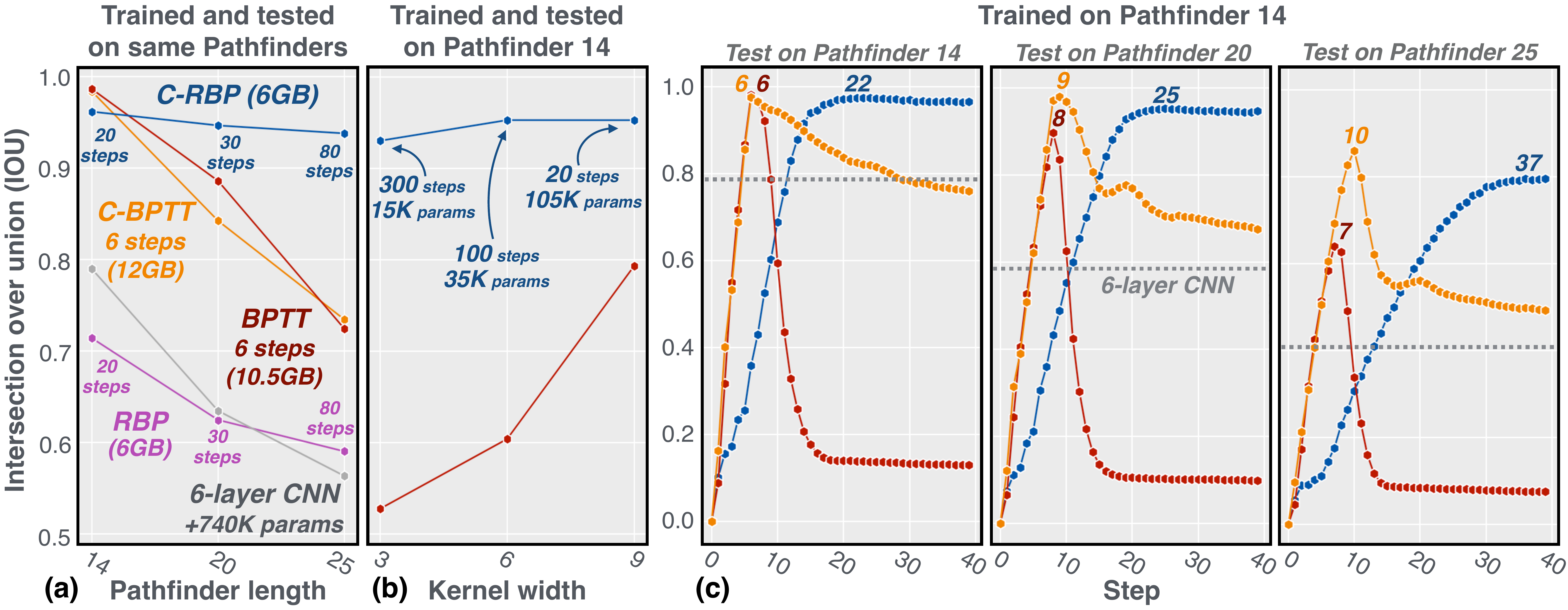}  
  \caption{Enforcing contraction in recurrent CNNs improves their performance, parameter efficiency, and enables our constant-memory \textcolor{CRBP}{C-RBP} learning algorithm. \textbf{(a)} hGRU models were trained and tested on different versions of \textit{Pathfinder}. Only the version trained with \textcolor{CRBP}{C-RBP} maintained high performance across the three datasets. \textbf{(b)} \textcolor{CRBP}{C-RBP} models can rely on recurrent processing rather than spatially broad kernels to solve long-range spatial dependencies. \textcolor{BPTT}{BPTT}-trained models cannot practically do this due to their linear memory complexity. \textbf{(c)} LCP improves the stability of hGRU dynamics and, as a result, the generalization of learned visual routines for contour integration. Models were trained on Pathfinder-14, and tested on all three Pathfinder datasets. hGRUs trained with \textcolor{CRBP}{C-RBP} and \textcolor{CBPTT}{C-BPTT} generalized far better than a version trained with \textcolor{BPTT}{BPTT} or a 6-layer CNN control. Numbers above each curve denote the max-performing step.}\label{fig:pf_results}\vspace{-4mm}
\end{center}\end{figure}


\subsection{Stable and expressive recurrent CNNs with constant memory complexity}Returning to \textit{Pathfinder}, we tested whether the LCP stabilizes the dynamics of models trained with RBP or BPTT. LCP consists of a single hyperparameter, $\lambda \in [0,1)$, which mediates the degree of contraction that is imposed. As $\lambda \to 0^+$, the local contraction at a fixed point becomes stronger leading to increased stability and reduced expressiveness. In general we find that $\lambda$ is task dependent, but on \textit{Pathfinder}, models performed well over many settings. We set $\lambda=0.9$ on the experiments reported here, and adopt the methods described in \S\ref{sec:background}.      

\paragraph{Learning a task-optimal trade-off between stability and expressivity.}
A BPTT-trained hGRU constrained with LCP performed as well on Pathfinder-14 as one trained with BPTT with no constraint (0.98 \textit{IoU} for both). However, a state space analysis of this ``contractor-BPTT'' (C-BPTT) hGRU revealed stable contractor dynamics, unlike the BPTT-trained hGRU (Fig.~\ref{fig:state_space}b).
We took this success as evidence that hGRUs trained with LCP satisfy the RBP \textit{constraint qualifications} test. We validated this hypothesis by training a 20-step hGRU with RBP \emph{and} LCP, which we refer to as ``contractor-RBP'' (C-RBP). This model performs nearly as well as both BPTT and C-BPTT trained hGRUs (0.95 \textit{IoU}), despite using approximately half the memory of either. The C-RBP hGRU also converged to a steady state that maintained task information (Fig.~\ref{fig:state_space}b; $t=T$), and like C-BPTT, the distance between its task-optimized hidden state $t=N$ and steady state $t=T$ was smaller than the BPTT hGRU. Pairwise testing of these distances with 2-sample Kolmogorov–Smirnov (KS) tests revealed that the BPTT hGRU diverged from $t=N$ to $t=T$ significantly more ($\mu=43.55$, $\sigma=0.53$, $p < 0.001$) than either C-BPTT or C-RBP hGRUs ($\mu=1.96$, $\sigma=2.73$ and $\mu=0.11$, $\sigma=0.18$, respectively; Fig.~\ref{fig:state_space}c). We repeated these experiments for convLSTMs and found similar results, including C-RBP improving its performance on \textit{Pathfinder} (Fig.~S8). C-RBP therefore achieves our main goal of constant-memory training with performance on par with BPTT. However, in the following experiments, we demonstrate several qualities that make C-RBP preferable to BPTT.

\paragraph{C-RBP can solve \textit{Pathfinder} tasks that BPTT cannot.} We tested how recurrent CNNs performed on harder versions of \textit{Pathfinder}, with 20- and 25-dash target paths, which forces them to learn longer-range spatial dependencies (Fig.~S2). The linear memory complexity of BPTT caps the number of steps that can fit into memory, leading to poor performance on these datasets (Fig.~\ref{fig:pf_results}a; BPTT and C-BPTT performance). Since C-RBP faces no such memory bottleneck, we were able to train 80-step C-RBP models to solve all versions of \textit{Pathfinder}, achieving $>0.90$ \textit{IoU} on each dataset and vastly outperforming other models, including BPTT/RBP-trained hGRUs/convLSTMs, and the 6-layer CNN control model.
\paragraph{C-RBP models can achieve better parameter efficiency through extensive recurrent processing.} \textit{Pathfinder} is a test of contour integration, a visual routine for chaining arbitrarily small receptive fields together along the extent of the target contour. The instantaneous receptive field sizes of the recurrent CNNs used here is a function of convolutional kernel sizes. Thus, we hypothesized that models with small kernels could solve \textit{Pathfinder} if they had sufficient steps of recurrent processing. Indeed, we found that hGRUs trained with C-RBP for 300 steps but only $3\times3$ kernels could solve Pathfinder-14 nearly as well as the baseline BPTT-trained hGRU with $15\times15$ kernels (Fig.~\ref{fig:pf_results}b).
\paragraph{Stable dynamics improve the out-of-distribution generalization of visual routines.} Visual routines such as contour integration derive long-range structure from repeated applications of a local grouping rule over steps of processing. This should lead to strong generalization on \textit{Pathfinder}, even when testing out-of-distribution, or between datasets. For instance, increasing target contour length in \textit{Pathfinder} affects the reaction time, but not accuracy, of human observers because they can solve the task through incremental grouping~\cite{Kim2020-yw,Houtkamp2010-wi}. If recurrent CNNs are learning similarly robust visual routines, a model trained on Pathfinder-14 should generalize to Pathfinder-20 and Pathfinder-25. To test this, we took hGRUs trained with BPTT, C-BPTT, or C-RBP on Pathfinder-14, and measured their performance on all versions of \textit{Pathfinder} (Fig.~\ref{fig:pf_results}c). Both C-BPTT and C-RBP trained hGRUs outperformed the BPTT-trained hGRU and feedforward control on each dataset, indicating that stable dynamics cause visual routines to generalize better. We found a similar improvement for C-RBP trained convLSTMs (Fig.~S8). BPTT-alternatives were not as effective as C-BPTT or C-RBP (Fig.~S8).
\begin{figure}
\begin{center}  
  \includegraphics[width=1\linewidth]{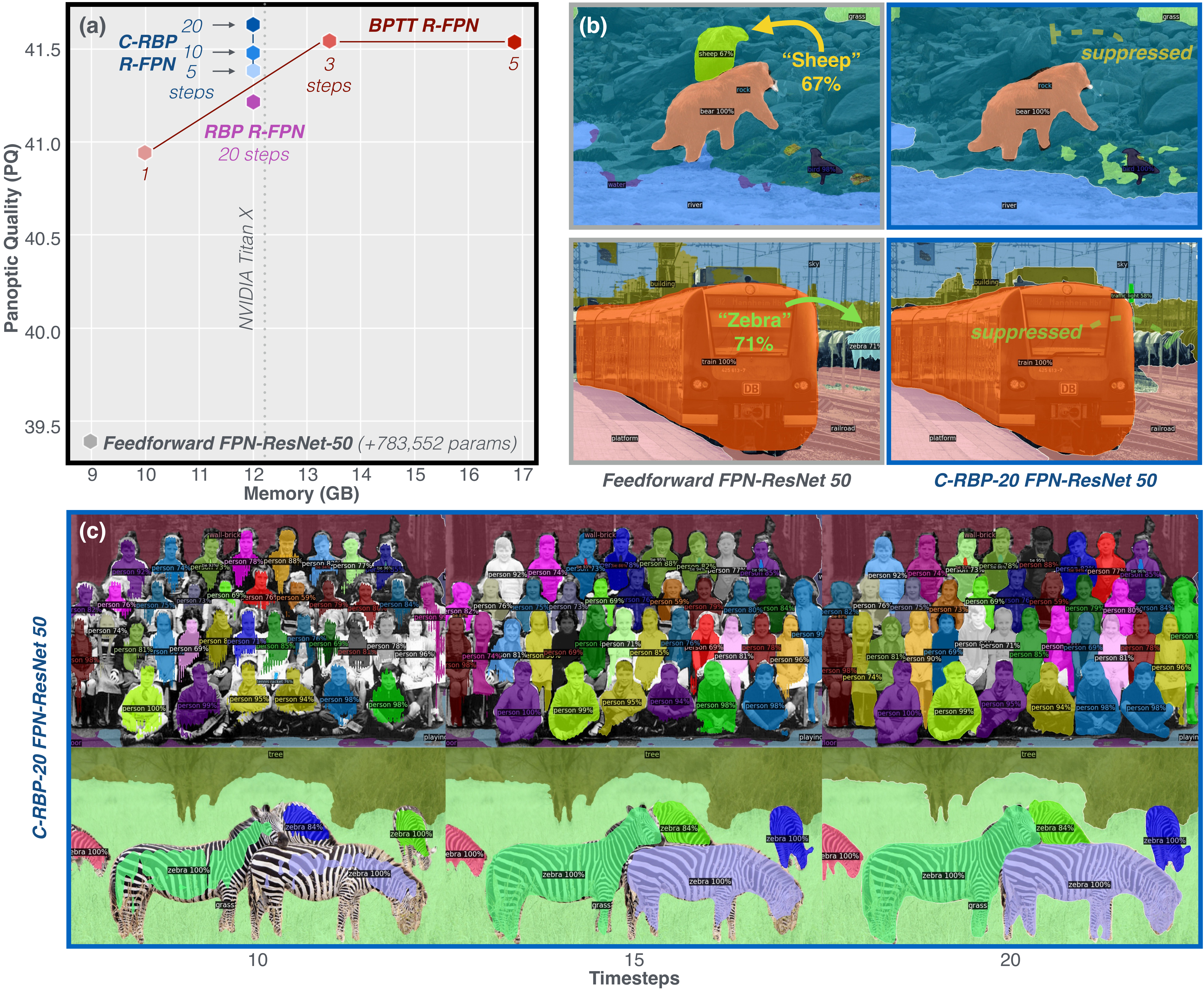}
  \caption{\textcolor{CRBP}{C-RBP} trained recurrent vision models outperform the feedforward standard on MS-COCO Panoptic Segmentation despite using nearly 800K fewer parameters. \textbf{(a)} Performance of our recurrent FPN-ResNet 50 trained with \textcolor{CRBP}{C-RBP} improves when trained with more steps of processing, despite remaining constant in its memory footprint. \textbf{(b)} Recurrent processing refines instance segmentations and controls false detections exhibited by the FPN-ResNet 50 (additional examples in SI). \textbf{(c)} Panoptic segmentation timecourses for an FPN-ResNet 50 trained with \textcolor{CRBP}{C-RBP} for 20 steps.}\label{fig:r50}\vspace{-4mm}
\end{center}\end{figure}
\subsection{Microsoft COCO Panoptic Segmentation}\label{sec:COCO}
Our main motivation in this paper is to understand -- and resolve -- the memory bottleneck faced by recurrent CNNs for large-scale computer vision challenges. To date, there have not been competitive recurrent solutions to the MS-COCO Panoptic Segmentation Challenge, a difficult task which involves recognizing the semantic category and/or object instance occupying every pixel in an image~\cite{Kirillov2019-og}. However, we will show that the stable dynamics and constant memory efficiency of C-RBP allows us to construct recurrent models that outperform the feedforward standards.

\paragraph{Methods}The leading approach to Panoptic Segmentation is an FPN Mask-RCNN~\cite{Kirillov2019-og} with a ResNet-50 or ResNet-101 layer backbone (refered to hereafter as FPN-ResNet-50 and FPN-ResNet-101). We developed recurrent extensions of these models, where we replaced the 4-layer Mask-RCNN head with an hGRU containing $3\times3$ kernels (SI \S3.4). We call these models R-FPN-ResNet-50 and R-FPN-ResNet-101. MS-COCO Panoptic Segmentation involves two separate objectives: (\textbf{i}) instance segmentation of the discrete and enumerable ``things'' in an image, such as people and animals; and (\textbf{ii}) semantic segmentation of the remaining background elements in scenes, or ``stuff'' as it is known in the challenge. Models were trained to optimize both of these objectives with SGD+Momentum, a learning rate of $5\text{e-}2$, and batches of 40 images across 24GB NVIDIA GTX GPUs (10 total). Model predictions on the COCO validation set were scored with Panoptic Quality (PQ), which is the product of metrics for semantic (\textit{IoU}) and instance ($F_1$ score) segmentation~\cite{Kirillov2019-og}. Note that differences between models in PQ scores can be interpreted as a percent-improvement on instance and semantic segmentation for the total image. Recurrent models trained with LCP used $\lambda=0.9$; training failed with higher values of this hyperparameter.
\paragraph{Results}After replicating benchmark performance of an FPN-ResNet-50 ($39.4 PQ$, 9GB memory; \url{https://github.com/crbp}), we evaluated our R-FPN models (Fig.~\ref{fig:r50}a). We first trained versions of the R-FPN-ResNet-50 with BPTT, which used nearly 17GB of memory for 5-steps of processing. BPTT performance increased with recurrence until it plateaued at 3-steps ($41.5 PQ$ for both 3- and 5-step models). Next, we trained an R-FPN-ResNet-50 with RBP for 20 steps, and found that this model ($41.22 PQ$, 12GB) performed better than the 1-step R-FPN-ResNet-50, but worse than a 2-step BPTT-trained model ($41.45 PQ$). R-FPN-ResNet-50 models improved when they were trained with C-RBP. A 5-step model trained with C-RBP ($41.39$, 12GB) outperformed a 20-step model trained with RBP; a 10-step C-RBP model was similar ($41.48 PQ$, 12GB) to the 3- and 5-step BPTT models; and a 20-step C-RBP model performed best ($41.63 PQ$, 12GB). Importantly, each of these R-FPN-ResNet-50 models was more accurate than the feedforward FPN-ResNet-50 despite using 783,552 \textit{fewer} parameters. We also found that a 20-step C-RBP R-FPN-ResNet-101 outperformed the feedforward FPN-ResNet-101 and BPTT-trained R-FPN-ResNet-101 models (Fig.~S10). 

\begin{figure}
\begin{center}  
  \includegraphics[width=0.99\linewidth]{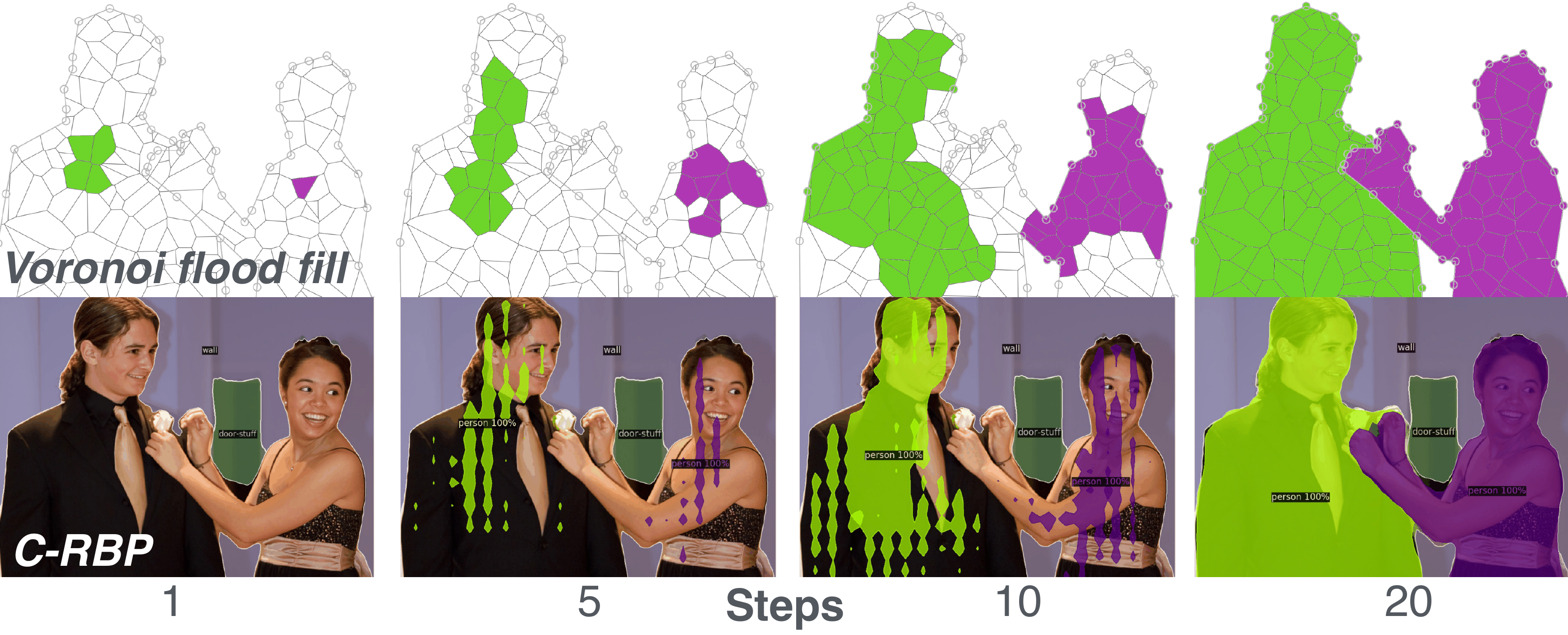}
  \caption{\textcolor{CRBP}{C-RBP}-optimized models for Panoptic segmentation learn to implement a routine that resembles a flood-fill algorithm, despite no explicit supervision to do so. \textbf{Top} Flood-filling in a Voronoi-tesselation of object masks. \textbf{Bottom} The routine learned by a \textcolor{CRBP}{C-RBP} R-FPN-ResNet-50.}\label{fig:floodfill} \vspace{-6mm}
\end{center}\end{figure}

There are two key qualitative improvements for the 20-step C-RBP R-FPN-ResNet-50 over the standard feedforward model. First, it suppressed false detections of the feedforward model, such as an over-reliance on texture for object recognition (Fig.~\ref{fig:r50}b; other examples in Fig.~S11-12). Second, and most surprisingly, the C-RBP model learned to solve the task by ``flood-filling'' \textit{despite no explicit constraints to do so} (Fig.~\ref{fig:r50}c). There is extensive work in the brain sciences suggesting that human and primate visual systems rely on a similar algorithm for object segmentation, classically referred to as ``coloring''~\cite{Ullman1984-gz,Jeurissen2016-dn}. We validated the extent to which the C-RBP R-FPN-ResNet-50 discovered a ``flood-filling'' segmentation routine by tessellating the ground-truth object masks in images, then applying a standard flood-filling algorithm to the resulting map. Both the C-RBP R-FPN-ResNet-50 and this idealized flood-filling model exhibited similar segmentation strategies (Fig.~\ref{fig:floodfill}).

\subsection{Related work}\label{si_sec:related_work}
\paragraph{Memory efficient learning algorithms} BPTT's memory bottleneck has inspired many efficient alternatives. A popular example is truncated back-propagation through time (TBPTT), which improves memory efficiency with a shorter time horizon for credit assignment. There are also heuristics for overcoming the BPTT memory bottleneck by swapping memory between GPU and CPU during training~\cite{Dean2012-qa,Rhu2016-xt}, or ``gradient checkpointing'', and recomputing a number of intermediate activities during the backwards pass~\cite{Dauvergne2006-yq,Chen2016-cg,Gruslys2016-zp}. Neural ODEs can train neural networks with continuous-time dynamics by optimizing through black box ODE solvers, but their computational demands and numerical issues result in poor performance on vision tasks (\cite{Dupont2019-we, Dupont2019-we}, see SI \S4.1 for an expanded discussion). Deep equilibrium models (DEQ) are another constant-memory complexity learning algorithm with good performance on sequence modeling, rivaling the state of the art~\cite{Bai2019-wd}. However, we found that training recurrent CNNs with DEQ is unstable, like with RBP (SI \S4.1). 
\paragraph{Stable RNNs} Recurrent model ``stability'' is typically analyzed \wrt training, where exploding or vanishing gradients are a byproduct of gradient descent~\cite{Bengio1994-bs,Miller2018-bu}. Models are stable if their Jacobians are upper-bounded by $\lambda$ (\ie it is $\lambda$-contractive). One way to achieve this is with architecture-specific constraints on model parameters, which has worked well for generative adversarial networks (GANs)~\cite{Yoshida2017-ni,Miyato2018-gp} and sequence modeling LSTMs~\cite{Miller2018-bu}. A simple alternative, is to directly penalize Jacobian approximations, as has been done for GANs~\cite{Gu2014-rk,Gulrajani2017-oa,Wei2018-rk}. However, these approximations have not been investigated for recurrent vision models or in the context of visual recognition.

\section{Limitations and Future Directions}
More work is needed to extend C-RBP to visual tasks other than the ones we analyzed here. For example, it is unclear whether C-RBP can extend to complex dynamical systems with limit cycles or strange attractors. It is possible that extensions are needed to transform C-RBP into a general method for balancing stability and expressivity when optimizing recurrent neural networks to explain arbitrary dynamical systems. We also did not examine complex recursive architectures where all recurrent layers are on the same dynamic ``clock'', such as in the $\gamma$-net of~\cite{Linsley2020-en}. We leave the details of this application to future work, which we anticipate will be especially impactful for computational neuroscience applications, where such hierarchical feedback models can explain biological data that feedforward models cannot~\cite{Linsley2020-en,Kim2020-yw,Kubilius2019-qr,Sinz2018-jp,Tang2018-we}. Another open problem is extending our formulation to spatiotemporal tasks, like action recognition, tracking, and closed-loop control. These tasks are defined by a time-varying feedforward drive, which presents novel optimization challenges.

\section{Conclusion}
There is compelling evidence from biological and artificial vision that the visual routines which support robust perception depend on recurrent processing~\cite{Roelfsema2000-op,Ullman1984-gz,Gilbert2013-hb}. Until now, the enormous cost of training recurrent CNNs with BPTT has made it difficult to test the hypothesis that these models can learn routines that will improve performance on large-scale computer vision challenges, which are dominated by high-capacity feedforward CNNs. C-RBP can alleviate this bottleneck, enabling constant-memory recurrent dynamics, more parameter efficient architectures, and systematic generalization to out-of-distribution datasets.

\section*{Broader Impact}
The development of artificial vision systems that can rival the accuracy and robustness of biological vision is a broadly useful endeavor for scientific research. Any such advances in artificial vision inevitably can lead to unethical applications. Because our methods and code are open sourced, our work is similarly exposed. However, we anticipate that our contributions to brain science make this risk worthwhile, and that our work will have a net positive impact on the broader scientific community.
\begin{ack}
We thank Michael Frank and Xaq Pitkow for insights which motivated this work; and Alexander Fengler, Remi Cadene, Mathieu Chalvidal, Andrea Alamia, and Amir Soltani for their feedback. Funding provided by ONR grant \#N00014-19-1-2029 and the ANR-3IA Artificial and Natural Intelligence Toulouse Institute. Additional support from the Brown University Carney Institute for Brain Science, Initiative for Computation in Brain and Mind, and Center for Computation and Visualization (CCV). 
\end{ack}

\bibliographystyle{splncs}
\bibliography{egbib}

\clearpage
\setcounter{figure}{0}
\makeatletter 
\renewcommand{\thefigure}{S\@arabic\c@figure}
\makeatother
\setcounter{table}{0}
\makeatletter 
\renewcommand{\thetable}{S\@arabic\c@table}
\renewcommand{\thefigure}{S\arabic{figure}}
\makeatother

\appendix

\section{Extended background}

\subsection{Backpropagation through time (BPTT)}
BPTT is the standard learning algorithm for optimizing recurrent parameters $w_F$ \wrt a loss $\mathcal{L}(\tilde{y}, y)$. It is implemented by replicating a dynamical system and accumulating its gradients over $N$ steps of processing (Eq.~\ref{eq:BPTT}, $K=0$). Given a recurrent function $F$ parameterized by $w_F$, which maintains a latent state $h_t$ for each time step $t$, BPTT is implemented by Eq.~\ref{eq:BPTT}.

\begin{equation}
\frac{\partial \mathcal{L}}{\partial w_F} = \frac{\partial \mathcal{L}}{\partial \tilde{y}}\frac{\partial \tilde{y}}{\partial h_T}\sum_{k=K}^{k=T-1}\left( \prod_{i=T}^{i=T-k}J_F(h_i)\right)\frac{\partial F}{\partial w_F}(x, h_{T-k}, w_F).\label{eq:BPTT}
\end{equation}
Here, $J_F(h_i)$ is the Jacobian of $F$ at $h$ on step $i$. Note that this algorithm stores every $h_t$ in memory during the forward pass, causing a memory footprint that linearly increases with steps.

\subsection{Lipschitz Coefficient Penalty}\label{si_sec:lcp}
We designed the \textit{Lipschitz Coefficient Penalty} (LCP) as a hyperparameter-agnostic regularization for forcing recurrent CNNs to learn contraction maps. As mentioned in the main text, LCP constrains the vector-Jacobian product $\left( \mathbf{1} \cdot J_{F, h^*}\right)_i < 1 \; \forall i$, where $i$ is a column index.
\begin{equation}
\lVert(\mathbf{1} \cdot J_{F, h^*}- \lambda)^+\rVert_2.
\end{equation}\label{si_eq:LCP}
\noindent Here, $(\cdot)^+$ denotes element-wise rectification and $\lambda \in [0,1)$ controls the degree of contraction in $F$. To derive LCP, we begin from the first-order Taylor expansion of $F(h)$, 
$$F(h) \approx F(\bar{h}) + J_{F, \bar{h}} \cdot (h-\bar{h})  + \cdots,$$
with which one can show:
\begin{equation}
\frac{\left\lVert F(h) - F(\bar{h}) \right\rVert_2}{\left\lVert h - \bar{h} \right\rVert_2} \approx \frac{\left\lVert  J_{F, \bar{h}} \cdot (h-\bar{h}) \right\rVert_2}{\left\lVert h - \bar{h} \right\rVert_2},
\label{eq:contractor_approx}
\end{equation}
Recalling the necessary condition for $F$ being a contractive map,
\begin{equation}
\left\lVert F(h) - F(\bar{h}) \right\rVert_2 < \lambda \left\lVert h - \bar{h} \right\rVert_2,
\label{eq:contractor_def}
\end{equation}
we can observe that the right hand side of Eq.~\ref{eq:contractor_approx} must be less than or equal to $\lambda \in [0,1)$ for any $h$ sufficiently close to $\bar{h}$. Thus, $F(\cdot)$ will be $\lambda$-contractive \emph{at least in the neighbourhood} of $\bar{h}$ if the LHS is forced to be less than $\lambda$. We accomplish this goal by explicitly regularizing $\bar{h} = h^*$ over the course of training:
\begin{equation}
\left\lVert J_{F, h^*} \cdot \hat{v}\right\rVert_2   < 1,
\end{equation}
for all unit vectors $\hat{v}$, which implies that the largest singular value of $J_{F, h^*}$ must be less than 1. This is equivalent to requiring $\left\lVert  J_{F, h^*}\right\rVert_2 < 1$. Note that $\lVert h - h^* \rVert_2$ may not necessarily be small for all $h$'s sampled along our trajectories, and so the Taylor approximation and hence Eq.~\eqref{eq:contractor_approx} may not hold. Nevertheless, in experiments on Pathfinder and Microsoft COCO our regularisation still yields reasonably stable convergence to fixed points.

Indeed, the matrix 2-norm is bounded from above and below by the 1-norm,
\begin{equation}
1/\sqrt{n} \left \lVert J_{F, h^*} \right \rVert_1 \leq \left \lVert J_{F, h^*} \right \rVert_2 \leq \sqrt{n} \left \lVert J_{F, h^*} \right \rVert_1,
\label{eq:1norm_bound}
\end{equation}
where 
\begin{equation}
\left \lVert J_{F, h^*} \right \rVert_1 = \max_i \sum_j \left\lvert \left(J_{F, h^*} \right)_{ij} \right\rvert,
\label{eq:1norm}
\end{equation}
and $n$ is the dimensionality of the Jacobian matrix. So if we can regularize and force $\sqrt{n} \left \lVert J_{F, h^*} \right \rVert_1$ to be below 1, then we can ensure that $F(\cdot)$ will be contractive. However, computing Jacobians of large matrices requires an enormous memory load, and it is far more efficient to compute vector-Jacobian products instead. We shall instead approximate the 1-norm by taking
\begin{equation}
\max_i  \left\lvert  \sum_j \left(J_{F, h^*} \right)_{ij} \right\rvert = \max_i \left( \mathbf{1} \cdot J_{F, h^*} \right)_i,
\end{equation}
where $\mathbf{1}$ denotes the row vector with all entries being 1. We note that in using this approximation, the right inequality in \eqref{eq:1norm_bound} ceases to be a strict upper bound, but we find that the approximation works well in practice due to large $n$.

We regularise model training with this approximation by requiring that $\left( \mathbf{1} \cdot J_{F, h^*}\right)_i < 1 \; \forall i$.  This yields our \textit{Lipschitz Coefficient Penalty} (LCP):
\begin{equation}
\lVert(\mathbf{1} \cdot J_{F, h^*} - \lambda)^+\rVert_2,
\end{equation}
which can be added to any task loss. Here, $(\cdot)^+$ denotes element-wise rectification and $\lambda \in [0,1)$ is a hand-selected constant controlling the bound on $\lVert J_{F, h^*} \lVert_2$ and hence the degree of contraction in $F$.



\section{Recurrent Back-prop}\label{si_sec:RBP}
We review the Recurrent Back-Prop (RBP) learning algorithm of~\cite{Almeida1987-me,Pineda1987-im}. Given a transition function $F$, which is parameterized by $w_F$ and applied to the static drive $x$, hidden state $h$ over $t \in \left\{ 1..N \right\}$ steps of processing: $h_{t+1} = F(x, h_t, w_F)$. We define a model readout,  $\tilde{y} = G(h_T, w_G)$, where $G$ is a task-optimized readout parameterized by $w_G$. We also introduce a loss $\mathcal{L}$ which yields a distance between predicted and ground-truth outputs. By differentiating the loss with respect to the weights, we obtain the gradients:
\begin{align}
\frac{\partial \mathcal{L}_\infty}{\partial w_G} &= \frac{\partial \mathcal{L}_\infty}{\partial y_\infty}\frac{\partial G(x; h^*, w_G)}{\partial w_G},\\
\frac{\partial \mathcal{L}_\infty}{\partial w_F} &= \frac{\partial \mathcal{L}_\infty}{\partial y_\infty}\frac{\partial y_\infty}{\partial h^*}\frac{\partial h^*}{\partial w_F}.
\end{align}
To obtain an expression for $\partial h^*/\partial w_F$, we first introduce the auxiliary function
\begin{equation}
\Psi(w_F, h) = h - F(x,h,w_F),
\end{equation}
where, at a fixed point, $\Psi(w_F,h^*) = 0$. Differentiating with respect to $w_F$, we obtain
\begin{align*}
\frac{\partial \Psi(w_F, h^*)}{\partial w_F} &= \frac{\partial h^*}{\partial w_F} - \frac{dF(x,h^*,w_F)}{d w_F}\\
&= \left(I - \frac{\partial F(x,h^*,w_F)}{\partial h^*} \right)\frac{\partial h^*}{\partial w_F} - \frac{\partial F(x,h^*,w_F)}{\partial w_F}\\
&= 0.
\end{align*}
Rearranging yields
\begin{equation}
\frac{\partial h^*}{\partial w_F} = \left(I - J_{F, h^*} \right)^{-1} \frac{\partial F(x,h^*,w_F)}{\partial w_F},
\end{equation}
where $J_{F, h^*}$ is the Jacobian matrix $\partial F(x,h^*,w_F) / \partial h^*$. The implicit function theorem guarantees the existence and uniqueness of a function mapping $w_F$ to $h^*$ and hence of $\partial h^* / \partial w_F$ provided (i) $\Psi(w_F, h)$ is continuously differentiable and (ii) $\left(I - J_{F, h^*} \right)$ is invertible. The main virtue of RBP is that the memory it requires to train the RNN is constant with respect to the granularity of dynamics (steps of processing).


\section{Recurrent vision models}\label{si_sec:hGRU}
\subsection{\textit{Pathfinder}}
\begin{wrapfigure}[14]{r}{0.6\textwidth}\vspace{-4mm}
  \centering
    \includegraphics[width=0.6\textwidth]{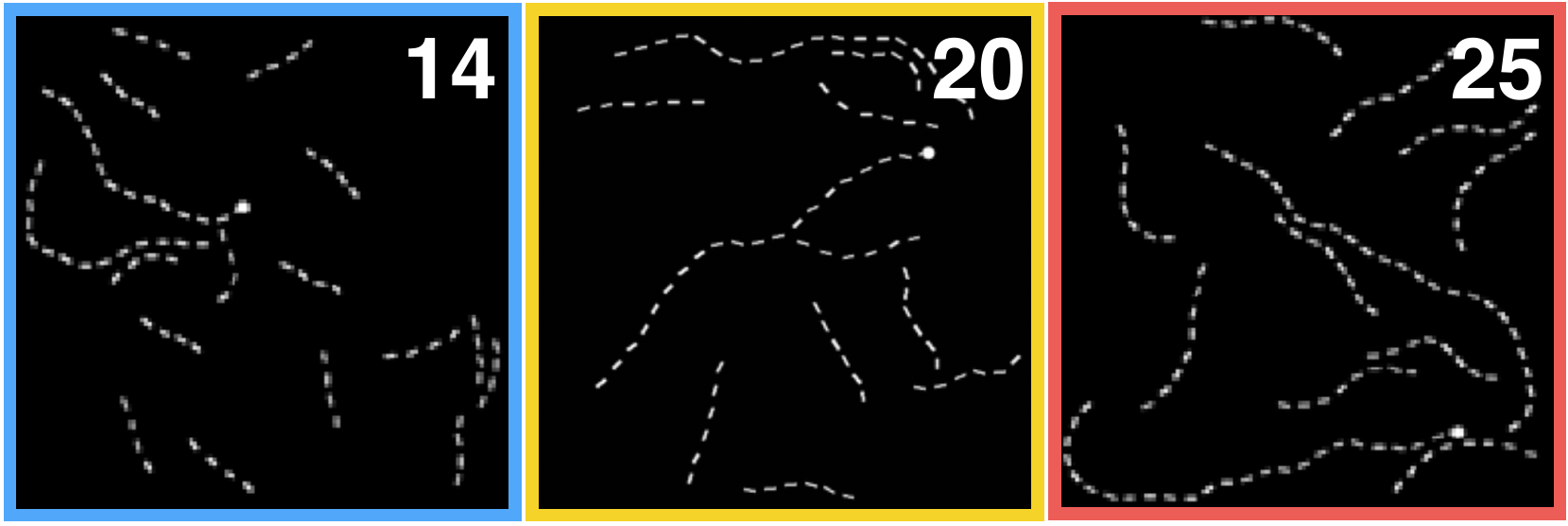}\vspace{-4mm}
  \caption{In our variation of the \textit{Pathfinder} challenge \cite{Linsley2018-ls,Kim2020-yw}, we ask observers to segment the contour connected to the white dot. Recurrent CNNs can easily solve it by learning to incrementally ``trace'' the target from end to end. Target contours are made up of 14-, 20-, or 25-dashes.}\label{fig_si:pfs}
\end{wrapfigure}
\paragraph{Horizontal Gated Recurrent Units}
The hGRU is a recurrent CNN, which when placed on top of a conventional feedforward convolutional layer, implements long-range nonlinear interactions between the feedforward layer's units~\cite{Linsley2018-ls}. These interactions take place over ``horizontal connections'' -- a concept from neuroscience, in which anatomical connections between nearby cortical neurons (in retinotopic space) are the substrate for complex recurrent processing. Tracing back to its origins as a neural circuit model, the hGRU distinguishes itself from other recurrent CNNs as having two distinct stages of processing with independent kernels in each. The first stage computes suppressive interactions, \ie a unit at location $(x,y)$ inhibits activity in a unit at location $(x+n, y)$, where $n$ is a spatial offset between these units. The second stage computes facilitative interactions, \ie a unit $(x,y)$ excites activity in a unit at location $(x, y+n)$. The hGRU is governed by the following equations:

\begin{align*}
\text{Stage 1:}\\
\textbf{A}^{S} &= U^{S} * \textbf{H}[t-1] && \text{\# Compute channel-wise selection}\\
\textbf{G}^{S} &= \mathit{sigmoid}(\textbf{A}^{S}) && \text{\# Compute suppression gate}\\
\textbf{C}^{S} &= \mathit{BN}(W^{S} * (\textbf{H}[t-1] \odot \textbf{G}^{S})) && \text{\# Compute suppression interactions}\\
\textbf{S}&=\lb \textbf{Z} - \lb (\alpha \textbf{H}[t-1] + \mu)\, \textbf{C}^{S} \rb_+ \rb_+, && \text{\# Additive and multiplicative suppression of } \textbf{Z}\\
\text{Stage 2:}\\
\textbf{G}^{F} &= \mathit{sigmoid}(U^{F} * \textbf{S}) && \text{\# Compute channel-wise recurrent updates}\\
\textbf{C}^{F} &= \mathit{BN}(W^{F} * \textbf{S}) && \text{\# Compute facilitation interactions}\\
\tilde{\textbf{H}}&=\lb\nu (\textbf{C}^{F} + \textbf{S}) + \omega (\textbf{C}^{F} * \textbf{S})\rb_+ && \text{\# Additive and multiplicative facilitation of } \textbf{S}\\
\textbf{H}[t]&= (1-\textbf{G}^{F}) \odot \textbf{H}[t-1] + \textbf{G}^{F} \odot \tilde{\textbf{H}} && \text{\# Update recurrent state} \label{hGRU} \\
\mbox{where } & 
\mathit { BN } ( \mathbf { R } ; {\delta} , {\nu} ) = {\nu} + {\delta} \odot \frac { \mathbf { R } - \widehat { \mathbb { E } } [ \mathbf { R } ] } { \sqrt { \widehat { \operatorname { Var } } [ \mathbf { R } ] + \eta } }.
\end{align*}
Here, $\textbf{H},\textbf{Z} \in \mathcal{R}^{\textit{X} \times \textit{Y} \times \textit{C}}$ are the hidden state and static drive from a preceding convolutional layer, respectively, with height/width/channels $X,Y,C$. Suppressive interactions in Stage 1 are computed with $W^{S} \in \mathbb{R}^{E \times E \times \textit{C} \times \textit{C}}$, and faciliative interactions in Stage 2 are computed with $W^{F} \in \mathbb{R}^{E \times E \times \textit{C} \times \textit{C}}$, where $E$ is the spatial extent of the horizontal connection kernel. In most of our experiments we set $E=15$, as in~\cite{Linsley2018-ls} (other kernel sizes were tested in our parameter efficiency analysis in Fig.~2b). The hGRU also contains gates to modulate input activity and interpolate the previous hidden state with the current step's state, $U^{S}, U^{F} \in \mathbb{R}^{1 \times 1 \times \textit{C} \times \textit{C}}$. Steps of processing are indexed by $t \in \left\{ 1..N \right\}$, and rectification using \textit{softplus} pointwise nonlinearities is denoted by $\lb\cdot\rb_+$, which ensures non-negativity in each stage, and hence, guarantees on suppression \vs facilitation. Lastly, we use batch normalization in the module to control exploding/vanishing gradients \cite{Cooijmans2017-bf}. This introduces two learned kernels, ${\delta}$, ${\nu} \in \mathbb { R } ^ { 1 \times 1 \times C }$, which control the scale and bias of normalization over input feature maps $\textbf{R}$, and are shared across steps of processing ($\eta$ is a small constant that protects divide-by-zero errors). As is standard in batch normalization, $\widehat { \mathbb { E } }$ and $\widehat { \operatorname { Var } }$ are estimated on-line during training.
\paragraph{Convolution LSTM} We use a standard implementation of convolutional LSTMs from~\cite{Lotter2016-qr}. These models used kernels with the same dimensions as those described above for the hGRU.
\subsection{State space analysis}\label{si_sec:state_space}
We analyzed the state space of recurrent models trained to solve \textit{Pathfinder}. This classic technique from dynamical systems has shown promise for analyzing computations of task-optimized recurrent neural networks~\cite{Maheswaranathan2019-pr}. Our approach to visualizing model state spaces involved the following steps: (\textbf{i}) Extract model hidden states for steps $t \in \left\{ 1..T \right\}$ elicited by a Pathfinder-14 image. (\textbf{ii}) Reduce hidden state dimensionality with a global average pool across spatial dimensions, yielding $C$-dimensional vectors. (\textbf{iii}) Fit a PCA using the $t \in \left\{ 1..N \right\}$ task-optimized steps of processing. (\textbf{iv}) Project all $t \in \left\{ 1..T  \right\}$ hidden states onto the extracted eigenvectors.
\subsection{Panoptic Segmentation}
As a proof-of-concept of C-RBP on large-scale computer vision challenges, we developed a straightforward recurrent extension to the leading feedforward approach to the MSCOCO Panoptic Segmentation challenge: the FPN-ResNet. This model uses a ResNet backbone (either 50- or 101-layers) pretrained on ImageNet, which passes its activities to a feature pyramid network (FPN; \cite{Lin2017-mj}). FPN activities are then sent to a linear readout for semantic segmentation (to identify the ``stuff'' in images) and a Mask-RCNN for instance segmentation (to identify and individuate the ``things''). We replace the Mask-RCNN head, which consists of 4-layers of convolutions and linear rectifications, with a single hGRU module (for both our version and the standard, activities from this stage are next upsampled, rectified, and linearly transformed into predictions). The hGRU that we used is slightly different than the one for the \textit{Pathfinder} challenge above. Batch normalization was replaced with group normalization~\cite{Wu2018-av}, which is standard for Panoptic segmentation. We also used a modification of the input gate, following~\cite{Linsley2020-en}, which was found to improve performance for natural image processing. Standard feedforward FPN-ResNets were approximately twice as fast to train as our 20-step C-RBP R-FPN-ResNets. Models with ResNet-50 backbones took between one and two days to train, whereas models with ResNet-101 backbones took two and four days to train.
\section{Extended discussion}
\subsection{Related work}
\paragraph{Recurrent vision models}
There are many successful applications of recurrent CNNS in computer vision, including object recognition, segmentation, and super-resolution tasks~\cite{Liang2015-sf,Kim2016-pg,Li2018-gx,Linsley2018-wx,Tang2018-we,George2017-ae,Lotter2016-qr}. These models often augment popular feedforward CNN architectures with local (within a layer) and/or long-range (between-layer) recurrent connections.

Others have found that augmenting recurrent CNNs with connectivity patterns or objective functions that are inspired by the anatomy or the physiology of the visual cortex can improve performance in visual reasoning, prediction, and recognition of occluded objects~\cite{Spoerer2017-ee,Lotter2016-qr,Zamir2016-lr,Linsley2018-wx,Wen2018-gf, Liao2016-ae, Tang2018-we,George2017-ae}.

\paragraph{Lipschitz constraints for stable \textit{training}}
There are many examples of using constraints on Lipschitz continuity to stabilize deep network training. This is especially popular for generative adversarial networks (GANs), where stability is enhanced by constraining the spectral norm of the weights~\cite{Yoshida2017-ni,Miyato2018-gp}, or Jacobians of each layer of the discriminator~\cite{Gu2014-rk}, or through Monte Carlo estimation of the discriminator's Jacobian~\cite{Gulrajani2017-oa,Wei2018-rk}. Penalizing the spectral norm of Jacobians can also yield better Auto Encoders~\cite{Rifai2011-wo}, and adversarial robustness in CNNs~\cite{Gu2014-rk}. In contrast to prior works on penalizing network Jacobians, in the current work we describe (i) an application to recurrent vision models, which (ii) enforces contractions only locally around equilibrium points (rather than globally across a hierarchy), which is a weaker constraint on model expressivity that still supports our key goal of stability during inference.


RNNs are notoriously challenging to train~\cite{Bengio1994-bs}, and the classic solution is to constrain Lipschitz continuity by introducing learnable gates~\cite{Hochreiter1997-gc, Miller2018-bu}. It should be emphasized that our models also take advantage of gates, and while these control vanishing and exploding gradients to stabilize training, they are not sufficient to yield contractive mappings. More generally, it has been found that stability is critical to train RNNs that can solve sequence modeling tasks~\cite{Simard1989-zk}, but that stability is less critical for BPTT~\cite{Miller2018-bu}. Other recent approaches induce stability via other architectural constraints, like weight orthogonalization via SVD~\cite{Zhang2018-dz}.

\paragraph{Deep networks as ODEs} Neural ODEs exploit the observation that the residual networks can be treated as a discrete-time ODE. By using black-box ODE solvers in the forward and backward passes of the network, this discretization can be taken towards zero~\cite{Chen2018-qz}. These models are trained with back-propagation through a latent trajectory derived from an adjoint system, giving them constant memory efficiency \wrt the granularity of dynamics, and they have shown promise in modeling continuous-time data and normalizing flows. However, Neural ODEs face several issues for computer vision applications. (\textit{i}) Neural ODEs are difficult to optimize because input-output mappings become arbitrarily complex over the course of training. (\textit{ii}) The adjoint method is slow to compute. (\textit{iii}) Neural ODEs require feature engineering to fit certain classes of non-linear data manifolds~\cite{Dupont2019-we}, and (\textit{iv}) they (along with the recent Augmented Neural ODEs~\cite{Dupont2019-we}) do not compare favorably to standard feedforward models on simple computer vision benchmarks like CIFAR. 

\paragraph{Deep Equilibrium Models}
A recent extension to RBP includes a Neumann-series computation to approximate the inverse of a dynamical system's Jacobian~\cite{Liao2018-tl}. Separately, deep equilibrium models (DEQ) use root-finding algorithms to exploit an identical formulation as in Eq.~3 to compute the steady state $h^*$~\cite{Bai2019-wd}. Both RBP and DEQ algorithms are effective for sequence modeling and meta-learning tasks, but have yet to be extended to vision. We attempted to train our hGRU models on Pathfinder 14 with DEQ, but it performed as poorly as the RBP-trained model, while using more GPU memory and taking longer to train.

\begin{figure}[t]
\begin{center}
  \includegraphics[width=.9\linewidth]{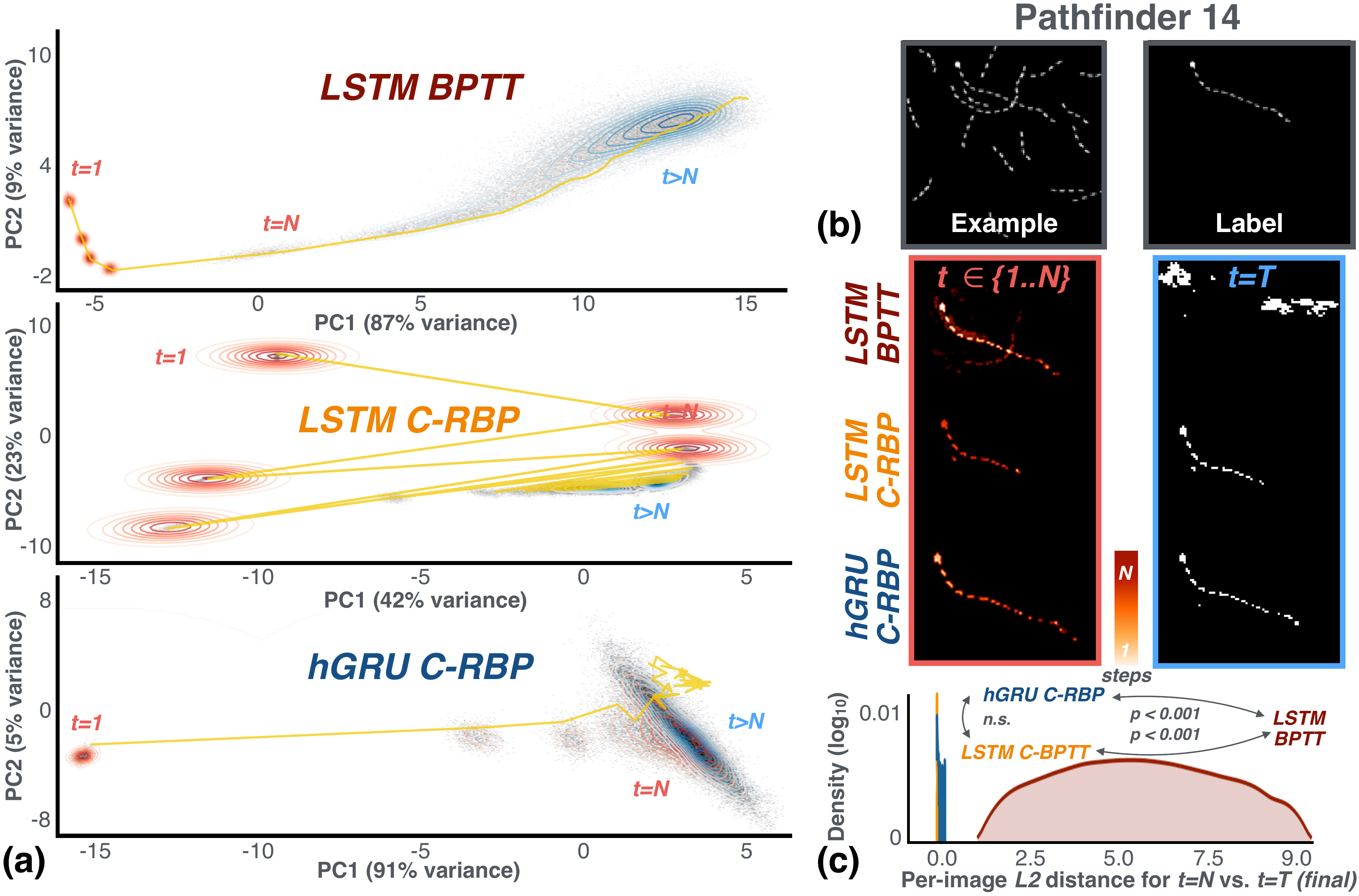}
  \caption{Convolutional LSTMs trained with (\textcolor{BPTT}{BPTT}) exhibit unstable dynamics, like the \textcolor{BPTT}{BPTT}-trained hGRUs examined in the main text. Once again, LCP corrects this pathology. \textbf{(a)} Visualization of convLSTM and hGRU state spaces following the state space method described in Section~\label{sec:si_state_space}. Here, the BPTT-LSTM was trained for 6 steps, the \textcolor{CBPTT}{C-RBP LSTM} for 60 steps, and the \textcolor{CRBP}{C-RBP hGRU} for 40 steps. Grey dots are the 2D-histogram of projected hidden states, red contours are hidden state densities up to the task-optimized $N$ steps, and blue contours are hidden state densities beyond that point ($t>N$). Exemplar dynamics for a single image are plotted in yellow. While dynamics of the \textcolor{BPTT}{BPTT} trained model diverge when $t>N$, models trained with LCP did not. \textbf{(b)} Model dynamics are reflected in their performance on Pathfinder-14 at $t=N$ and $t=T$ steps of processing. \textbf{(c)} Two-sample KS-tests indicate that the distance in state space between $t=N$ and $t=T$ hidden states is significantly greater for the BPTT-trained convLSTM than for either of the models trained with \textcolor{CRBP}{C-RBP} (n.s.$\ =\ $ not significant).}\label{fig_si:state_space_lstm}
\end{center}\end{figure}

\begin{figure}[t]
\begin{center}
  \includegraphics[width=1\linewidth]{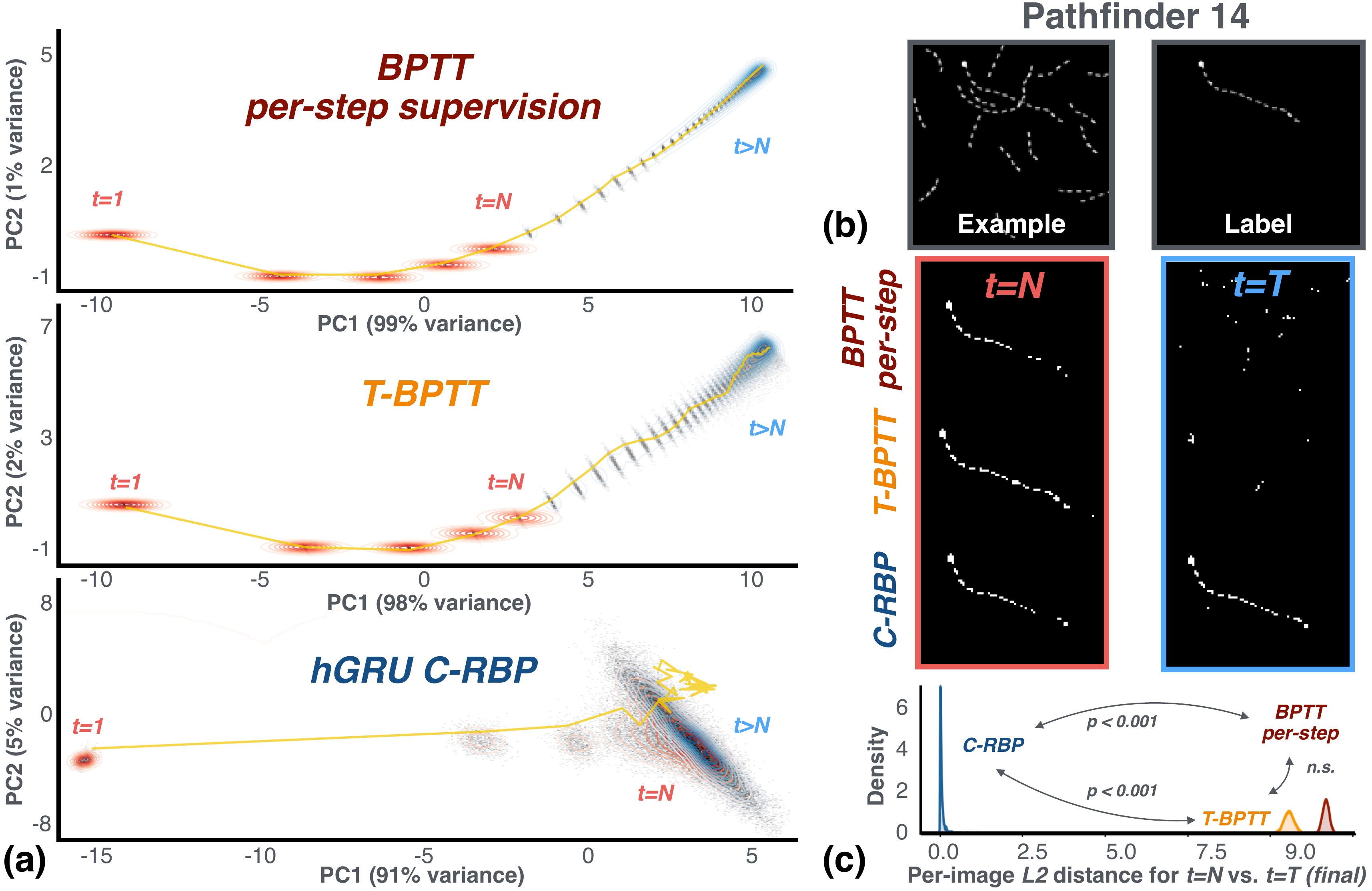}
  \caption{Additional state space analyses showed that alternatives to BPTT do not resolve the unstable dynamics we observed for recurrent CNNs. Here, \textcolor{BPTT}{BPTT per-step supervision} refers to a model which was optimized with a loss evaluated on each of its 6 steps of processing. \textcolor{CBPTT}{T-BPTT} refers to a model trained with truncated backprop, for which gradients were accumulated over 3 steps of its 6 steps of processing. \textbf{(a,b,c)} These BPTT alternatives train models with unstable dynamics, which forgot task information after the optimized $t=N$ steps of processing. The distances between $t=N$ and $t=T$ hidden states are significantly greater for hGRUs trained with these algorithms than for an hGRU trained with \textcolor{CRBP}{C-RBP} (n.s.$\ =\ $ not significant).}\label{fig_si:state_space_alt_bptt}
\end{center}\end{figure}

\begin{figure}[t]
\begin{center}
  \includegraphics[width=1\linewidth]{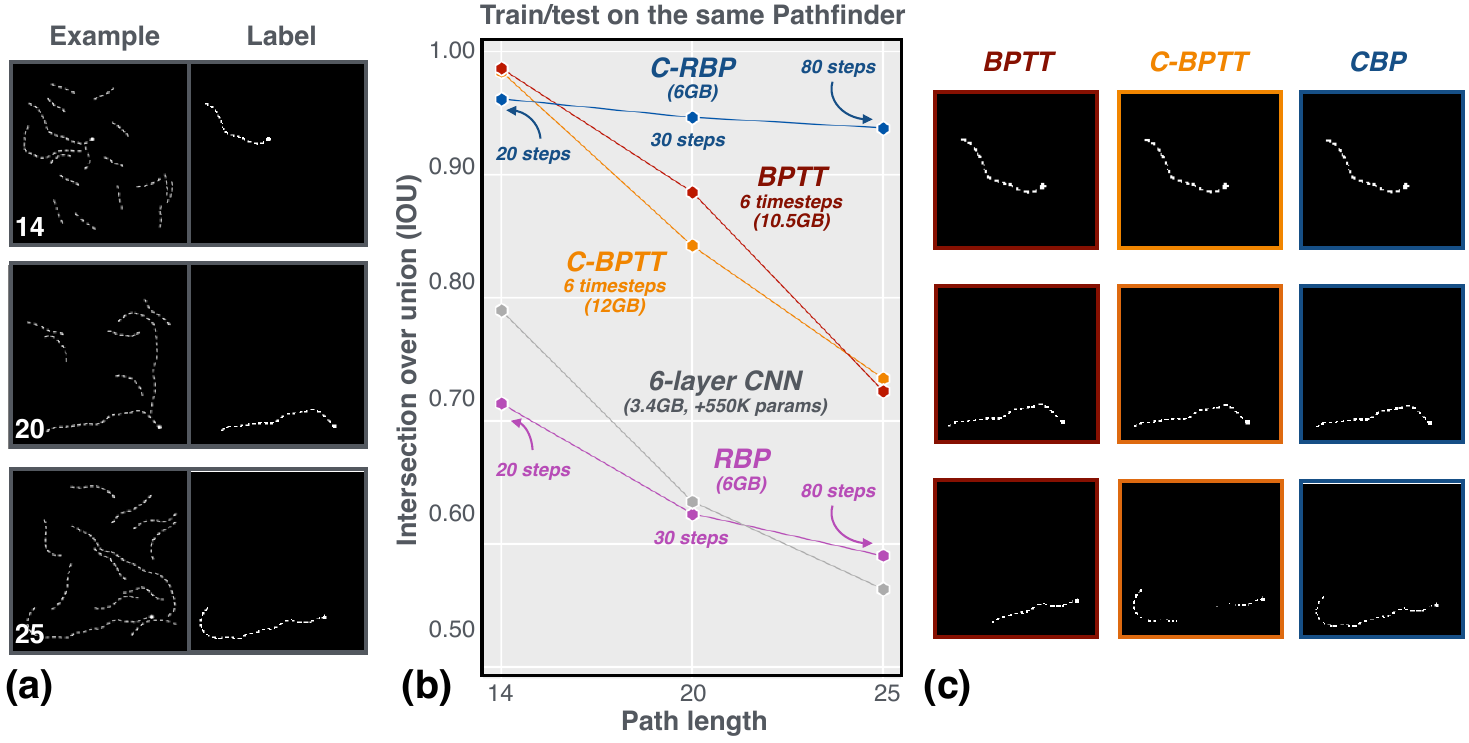}
  \caption{Exemplars from the (\textbf{a}) \textit{Pathfinder} challenge, along with (\textbf{b}) model performance on each of these datasets, and (\textbf{c}) predicted contours from hGRUs trained with the different algorithms.}\label{fig_si:train_test_same}
\end{center}\end{figure}

\begin{figure}[t]
\begin{center}
  \includegraphics[width=1\linewidth]{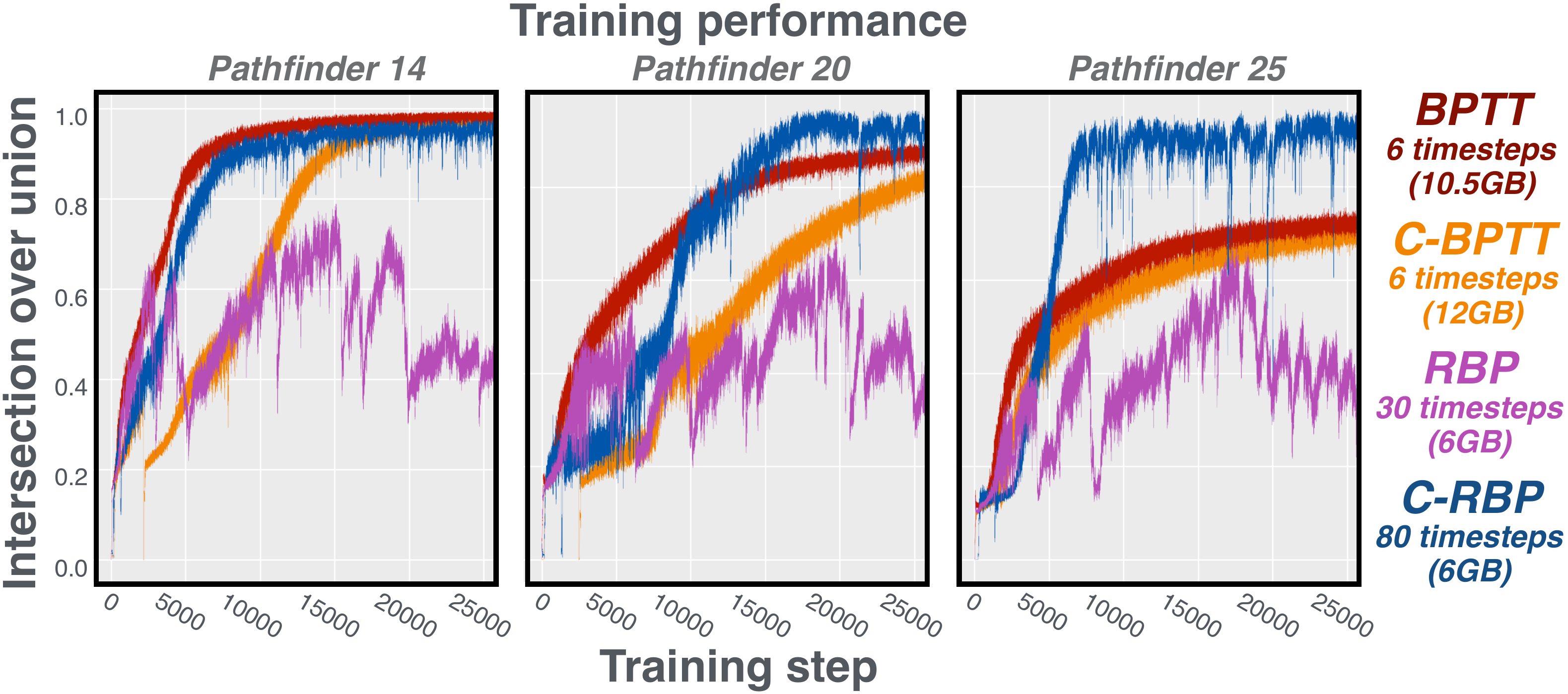}
  \caption{Performance of hGRUs during training on \textit{Pathfinder} challenge datasets. The \textcolor{RBP}{RBP}-trained model struggles to fit any dataset, unlike the models trained with \textcolor{BPTT}{BPTT}, \textcolor{CBPTT}{C-BPTT}, or \textcolor{CRBP}{C-RBP}.}\label{fig_si:training}
\end{center}\end{figure}

\begin{figure}[t]
\begin{center}
  \includegraphics[width=0.5\linewidth]{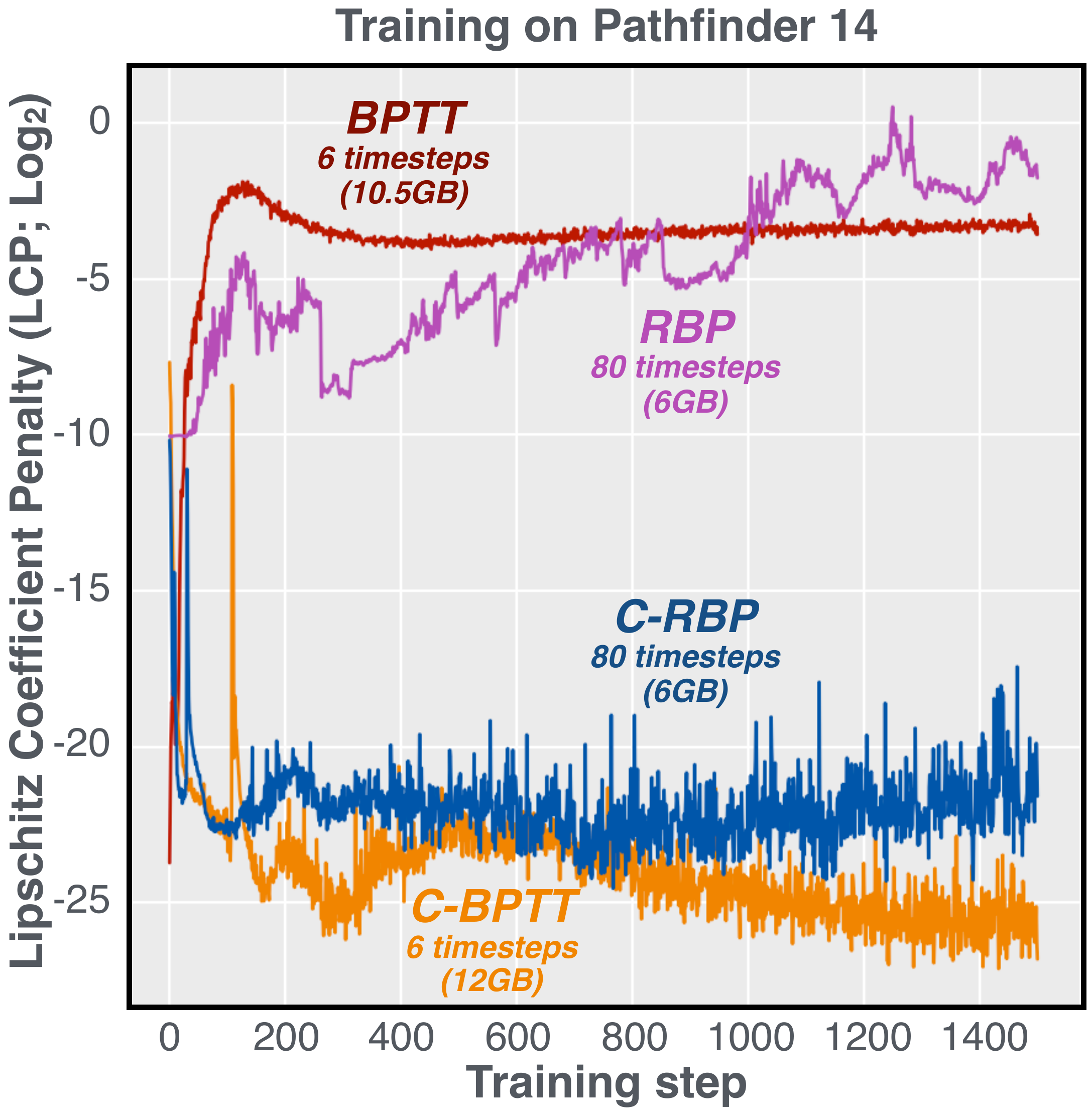}
  \caption{The value of our LCP (computed with Eq.~6 in the main text) over the course of training for models that minimize it (\textcolor{CRBP}{C-RBP}, \textcolor{CBPTT}{C-BPTT}) and models that do not (\textcolor{RBP}{RBP}, \textcolor{BPTT}{BPTT}). In other words, the magnitude of this correlates with the stability/instability of model dynamics.}\label{fig_si:penalty}
\end{center}\end{figure}


\begin{figure}[t]
\begin{center}
  \includegraphics[width=1\linewidth]{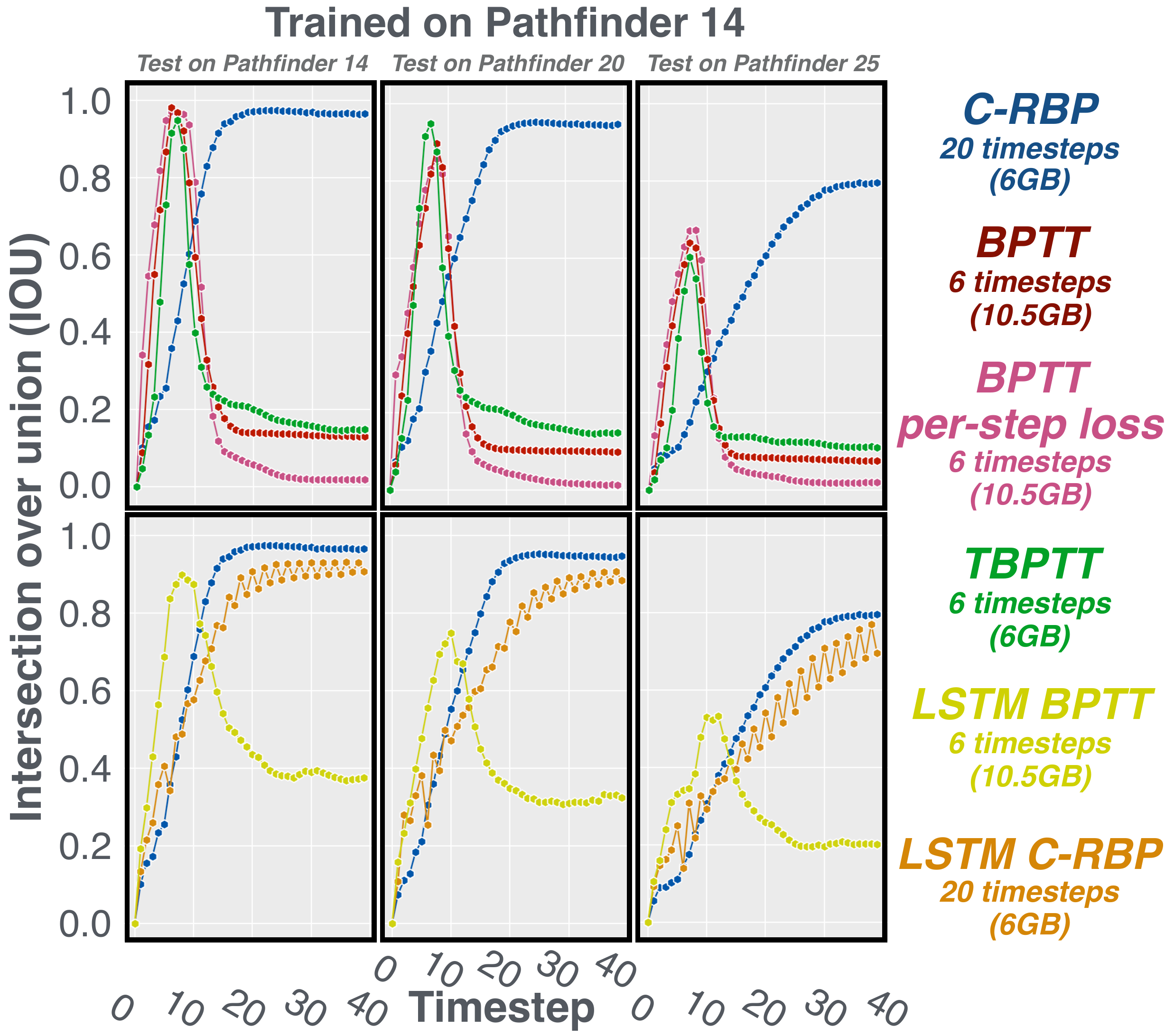}
  \caption{Generalization performance for hGRUs and convLSTMs trained with BPTT and alternatives to BPTT. Models were trained on Pathfinder 14 and tested on Pathfinder 14/20/25. For reference, performance of the hGRU trained with C-RBP is plotted in both rows. BPTT per-step loss means that a loss was computed on each of the 6 steps of hGRU training, and weights were optimized with BPTT. In contrast, a loss was only calculated on the final step for BPTT. TBPTT is truncated backprop through time, where gradients were computed over 3 steps of the 6 step dynamics. The LSTM trained with BPTT was trained for 6 steps, whereas the LSTM trained with C-RBP was trained for 60.}\label{fig_si:pf_alternative_perf}
\end{center}\end{figure}

\begin{figure}[t]
\begin{center}
  \includegraphics[width=.7\linewidth]{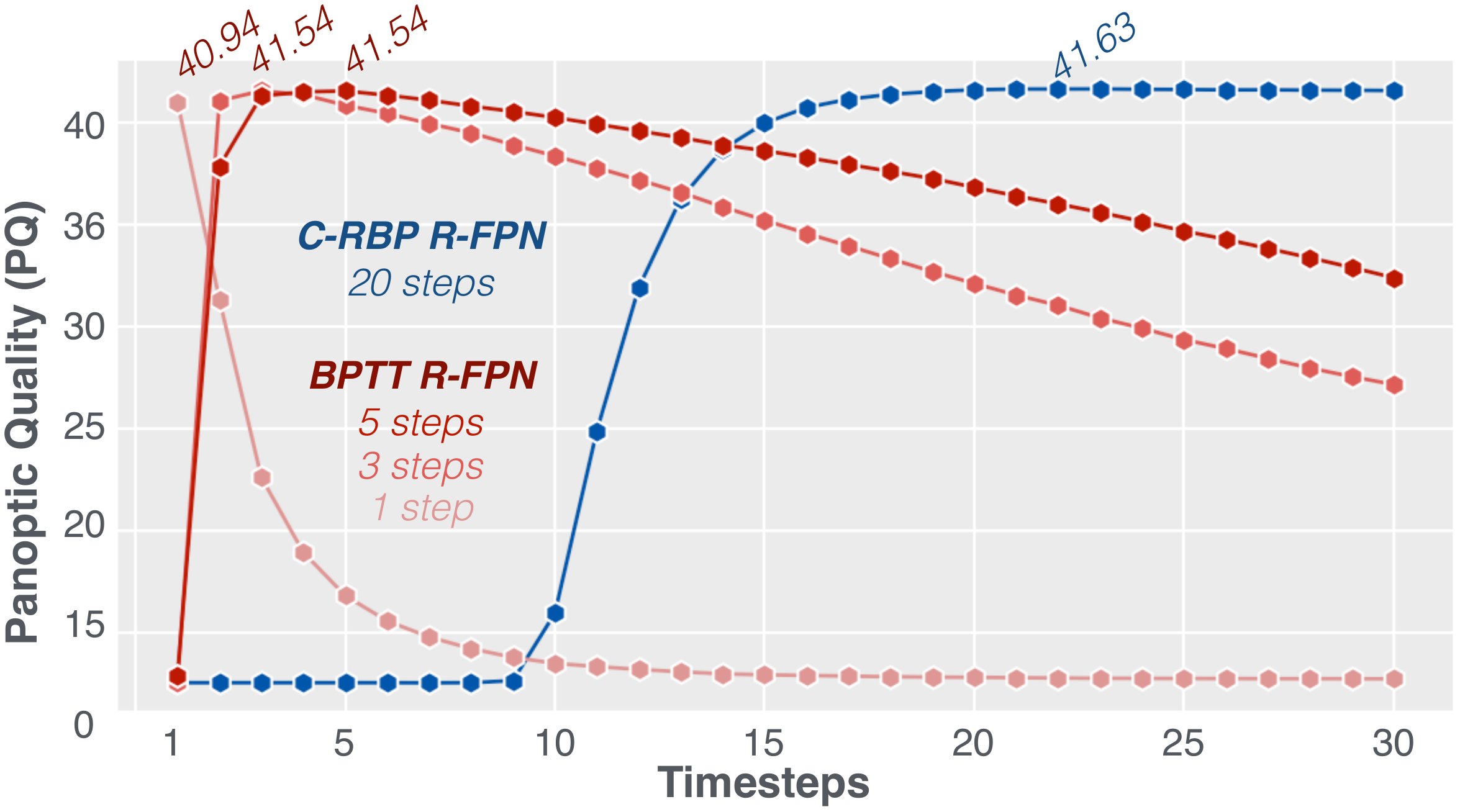}
  \caption{Recurrent model performance on MSCOCO Panoptic Segmentation. Performance was computed for each of 30 steps of processing for models trained with BPTT and C-RBP. The C-RBP models achieve better -- and more stable -- performance.}\label{fig_si:pq_timecourse}
\end{center}\end{figure}

\begin{figure}[t]
\begin{center}
  \includegraphics[width=0.5\linewidth]{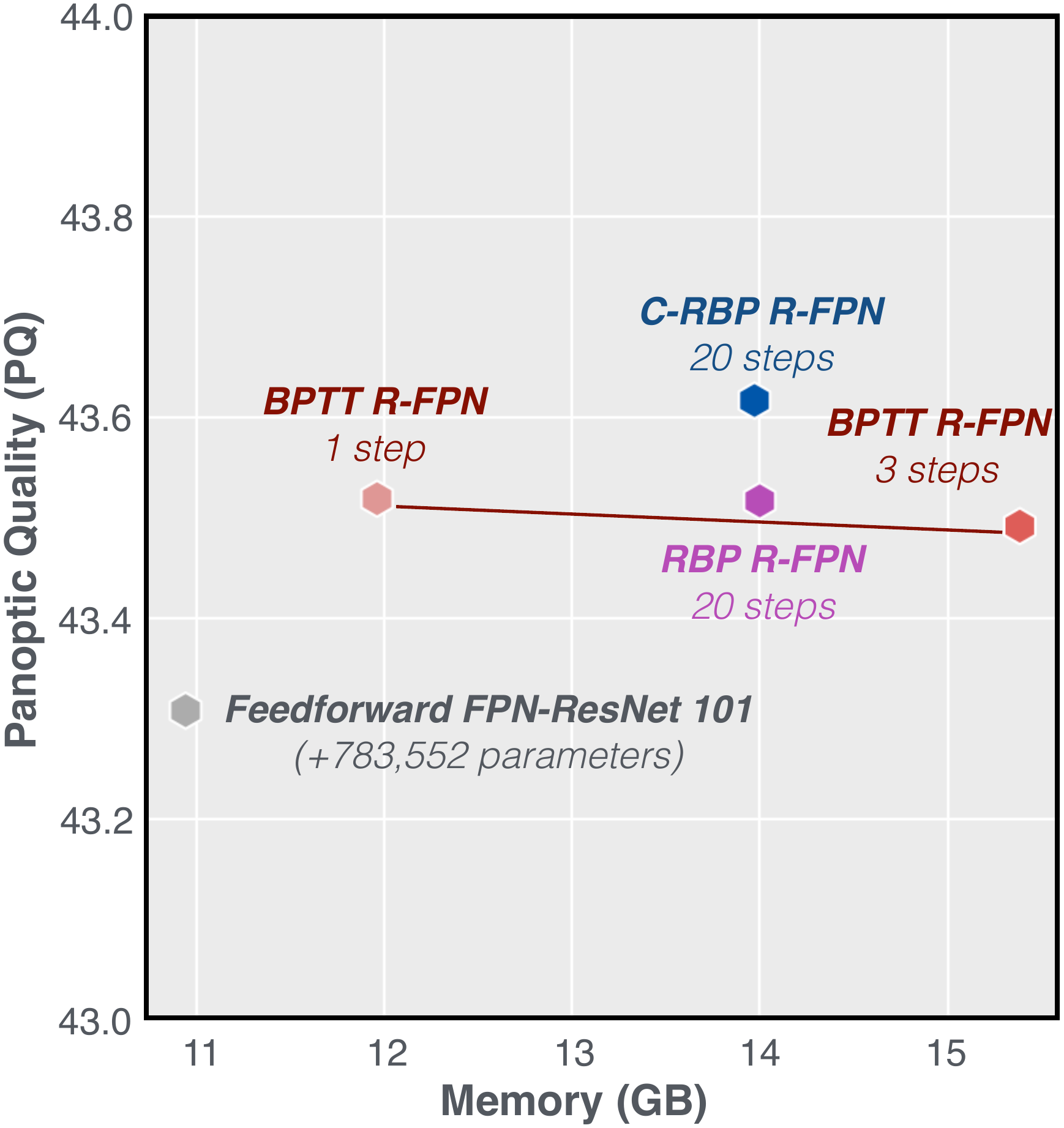}
  \caption{Our recurrent Panoptic Segmentation models also outperform the feedforward standard when both are given ResNet-101 backbones and the 3$\times$ training schedule (see \url{https://bit.ly/dtcon} for details on this training routine). The C-RBP model, trained for 20 steps, outperforms any other tested version of the model.}\label{fig_si:pq_101}
\end{center}\end{figure}

\clearpage
\begin{figure}[t]
\begin{center}
  \includegraphics[width=0.99\linewidth]{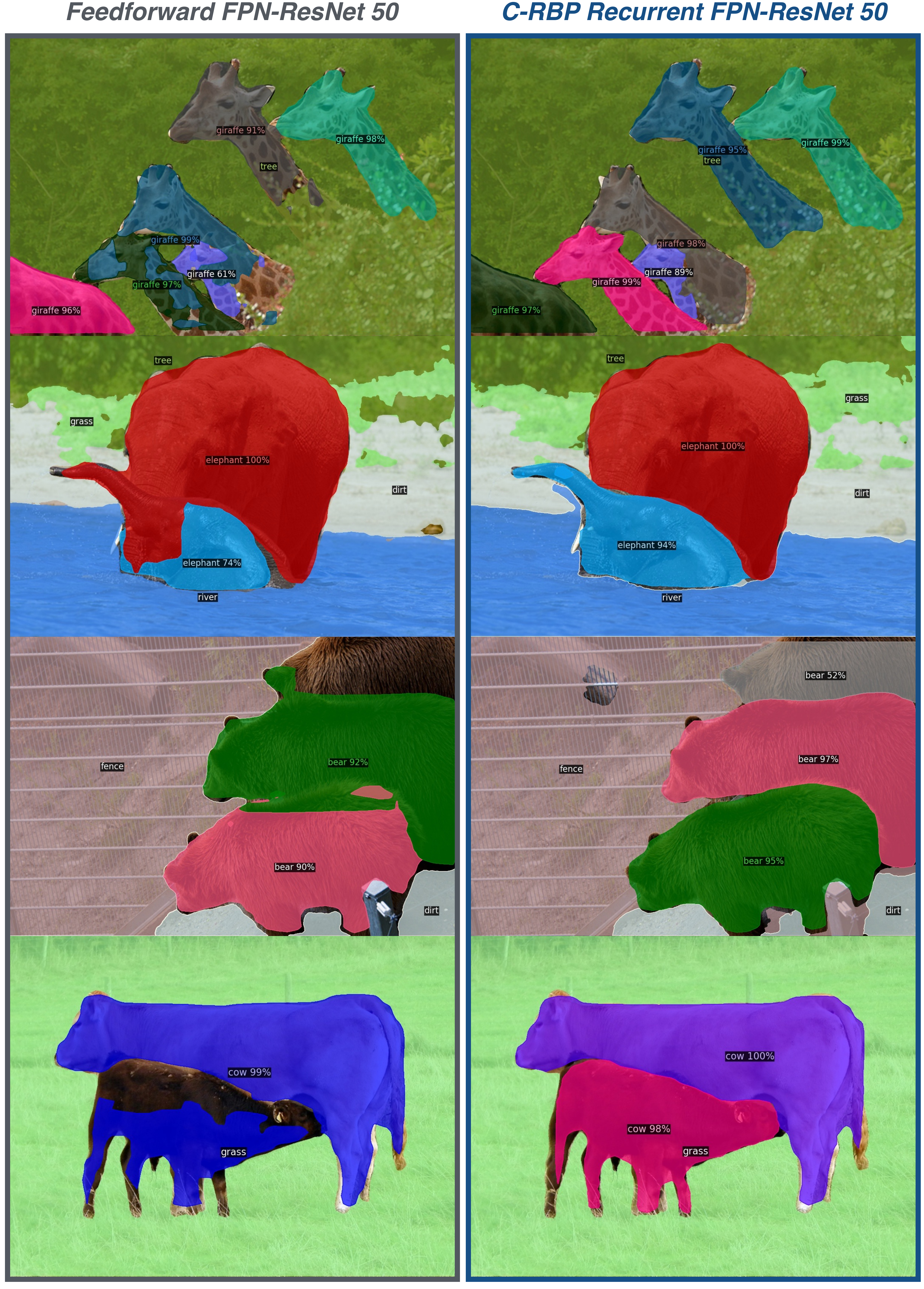}
  \caption{Panoptic predictions from the feedforward ResNet-50 FPN Mask-RCNN (left) and our recurrent version of the model trained with C-RBP (right).}\label{fig_si:extra_panoptic_0}
\end{center}\end{figure}
\clearpage
\begin{figure}[t]
\begin{center}
  \includegraphics[width=0.99\linewidth]{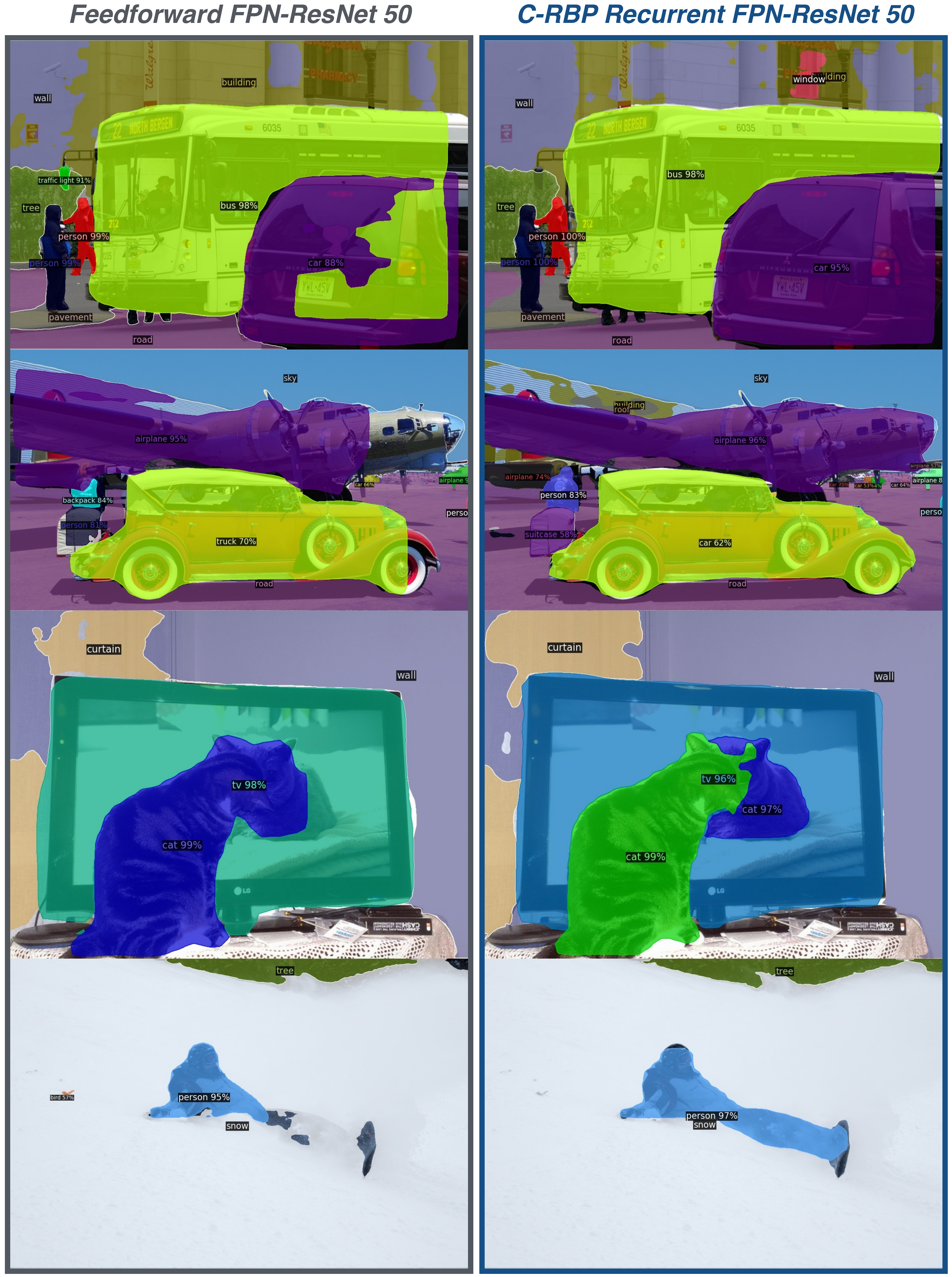}
  \caption{Panoptic predictions from the feedforward ResNet-50 FPN Mask-RCNN (left) and our recurrent version of the model trained with C-RBP (right).}\label{fig_si:extra_panoptic_1}
\end{center}\end{figure}

\end{document}